\documentclass{article}

\usepackage{arxiv}

\usepackage[utf8]{inputenc} 
\usepackage[T1]{fontenc}    
\usepackage{hyperref}       
\usepackage{url}            
\usepackage{booktabs}       
\usepackage{amsfonts}       
\usepackage{nicefrac}       
\usepackage{microtype}      
\usepackage{lipsum}		
\usepackage{graphicx}
\usepackage{doi}
\usepackage{tikz}
\usetikzlibrary{matrix,positioning}
\usepackage{xcolor}
 
\usepackage{color}
\usepackage{soul}
\usepackage{listings}
\usepackage{caption}
\usepackage{sansmath} 

\usepackage[skins,breakable]{tcolorbox}

\usepackage{multirow}
\usepackage{helvet} 
\usepackage{tikz}
\usetikzlibrary{shapes.geometric, arrows, arrows.meta, positioning,fit,backgrounds, shadows}
\usepackage{enumitem}

\tikzstyle{startstop} = [rectangle, rounded corners, minimum width=8.5cm, minimum height=1cm, text centered, text width=8cm, draw=black, fill=red!30]
\tikzstyle{result_rounded} = [rectangle, rounded corners, minimum width=8.5cm, minimum height=1cm, text centered, text width=8cm, draw=black, fill=yellow!30]

\tikzstyle{process} = [rectangle, minimum width=3cm, minimum height=1cm, text centered, text width=3cm, draw=black, fill=blue!30]
\tikzstyle{decision} = [diamond, minimum width=3cm, minimum height=1cm, text centered, text width=3.5cm, draw=black, fill=green!30]
\tikzstyle{graph} = [rectangle, minimum width=4.5cm, minimum height=1cm, text centered, text width=4cm, draw=black, fill=orange!30]

\tikzstyle{arrow} = [thick,->,>=stealth]
\tikzstyle{title} = [draw=none, fill=none, text centered, font=\large\bfseries]

\tikzstyle{arrow} = [thick,->,>=stealth]
\tikzstyle{data} = [rectangle, minimum width=8.5cm, minimum height=1cm, text centered,  text width=8cm, draw=black, fill=yellow!30]

\definecolor{salmon}{RGB}{250,128,114}
\definecolor{lightblue}{RGB}{173,216,230}
\definecolor{lightgreen}{RGB}{144,238,144}
\definecolor{yellow}{RGB}{255,255,153}
\definecolor{lightgray}{RGB}{211,211,211}
\definecolor{gray}{RGB}{169,169,169}

\tcbset{
  customboxsmalll/.style={
    width=0.95\textwidth,  
    fonttitle=\ttfamily\scriptsize, 
    colback=white,  
    colframe=black,  
    colbacktitle=blue!75!black, 
    coltitle=white,  
    boxrule=0.5pt,  
    rounded corners,  
    boxed title style={
        colback=blue!75!black, 
        colframe=black, 
        boxrule=0.5pt  
            before={\setlength{\parskip}{0pt} \setlength{\parindent}{-6pt}},  
    after={\setlength{\parskip}{0pt} \setlength{\parindent}{-6pt}},   
    halign=center,  
    before skip=-5pt,  
    after skip=-5pt,   
    innertop=-3pt,     
    innerbottom=-3pt,  
    valign=center,    
    }
    
  }
}

\tcbset{
  custombox/.style={
    width=512.18663pt,
    top=3pt,
    fonttitle=\bfseries\small\sffamily, 
    colback=white,
    colframe=black,
    colbacktitle=black,
    boxrule=0.5pt, 
    colback=white, 
    enhanced,
    rounded corners, 
    center,
    boxed title style={
            sharp corners,
            size=small,
            colframe=blue!75!black, 
        },
    attach boxed title to top left={yshift=-0.1in,xshift=0.15in},
    boxed title style={boxrule=0pt,colframe=white,},
    before=\setlength{\baselineskip}{-2pt}, 
    before upper=\setlength{\baselineskip}{8pt}, 
    after=\setlength{\baselineskip}{8pt}, 
  }
}

\newtcolorbox{LLMboxSmall}[2][]{customboxsmalll, title=#2,#1}

\newtcolorbox{LLMbox}[2][]{custombox, title=#2,#1}

\newcommand{\hlyellow}[1]{{\sethlcolor{white}\hl{#1}}}

\usepackage{amsmath}

\title{Graph-Aware Isomorphic Attention for Adaptive Dynamics in Transformers}

\author{ \href{https://orcid.org/0000-0002-4173-9659}{\includegraphics[scale=0.06]{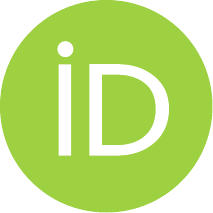}\hspace{1mm}Markus J. Buehler}\thanks{Corresponding author.} \\
	Laboratory for Atomistic and Molecular Mechanics (LAMM)\\
	MIT\\
	Cambridge, MA 02139, USA \\
	\texttt{mbuehler@MIT.EDU} \\
}

\renewcommand{\shorttitle}{\textit{arXiv} preprint}

\hypersetup{
pdftitle={Graph-Aware Isomorphic Attention for Adaptive Dynamics in Transformers
},
pdfsubject={q-bio.NC, q-bio.QM},
pdfauthor={Markus J. Buehler},
pdfkeywords={First keyword, Second keyword, More},
}

\begin{document}
\maketitle

\begin{abstract}
We present an approach to modifying Transformer architectures by integrating graph-aware relational reasoning into the attention mechanism, merging concepts from graph neural networks and language modeling. Building on the inherent connection between attention and graph theory, we reformulate the Transformer’s attention mechanism as a graph operation and propose Graph-Aware Isomorphic Attention. This method leverages advanced graph modeling strategies, including Graph Isomorphism Networks (GIN), to enrich the representation of relational structures. Our approach improves the model’s ability to capture complex dependencies and generalize across tasks, as evidenced by a reduced generalization gap and improved learning performance. We expand the concept of graph-aware attention to introduce Sparse-GIN-Attention, a fine-tuning approach that enhances the adaptability of pre-trained foundational models with minimal computational overhead, endowing them with graph-aware capabilities. We show that the Sparse-GIN-Attention framework leverages compositional principles from category theory to align relational reasoning with sparsified graph structures, while modeling hierarchical representation learning that bridges local interactions and global task objectives across diverse domains. Our results demonstrate that graph-aware attention mechanisms outperform traditional attention in both training efficiency and validation performance. These insights bridge graph theory and Transformer architectures and uncover latent graph-like structures within traditional attention mechanisms, offering a new lens through which Transformers can be optimized. By evolving Transformers as hierarchical GIN models, we reveal their implicit capacity for graph-level relational reasoning with profound implications for foundational model development and applications in bioinformatics, materials science, language modeling, and beyond, setting the stage for interpretable and generalizable modeling strategies.  
\end{abstract}


\section{Introduction}

The evolution of attention mechanisms in neural networks has significantly influenced the field of artificial intelligence and machine learning, from early work~\cite{hopfield1982neural,bahdanau2014neural,graves2014neural,hochreiter1997long, Vaswani2017AttentionNeed} to more recent developments~\cite{AlecRadfordImprovingPre-Training,Xue2021ByT5:Models,Jiang2023Mistral7Bb,Phi-2:Research,dubey2024llama3herdmodels,Brown2020LanguageLearners,buehler2024preflexorpreferencebasedrecursivelanguage,ghafarollahi2024sciagentsautomatingscientificdiscovery,qiu2024askshallgiventuring}. Originally vaguely inspired by cognitive processes in humans, attention mechanisms have become integral to modern neural network architectures like the Transformer~\cite{Vaswani2017AttentionNeed}. These mechanisms dynamically allocate computational resources to the most relevant parts of input data, optimizing the processing of information and reducing computational redundancy.

\begin{figure}
	\centering
	\includegraphics[width=1.\linewidth]{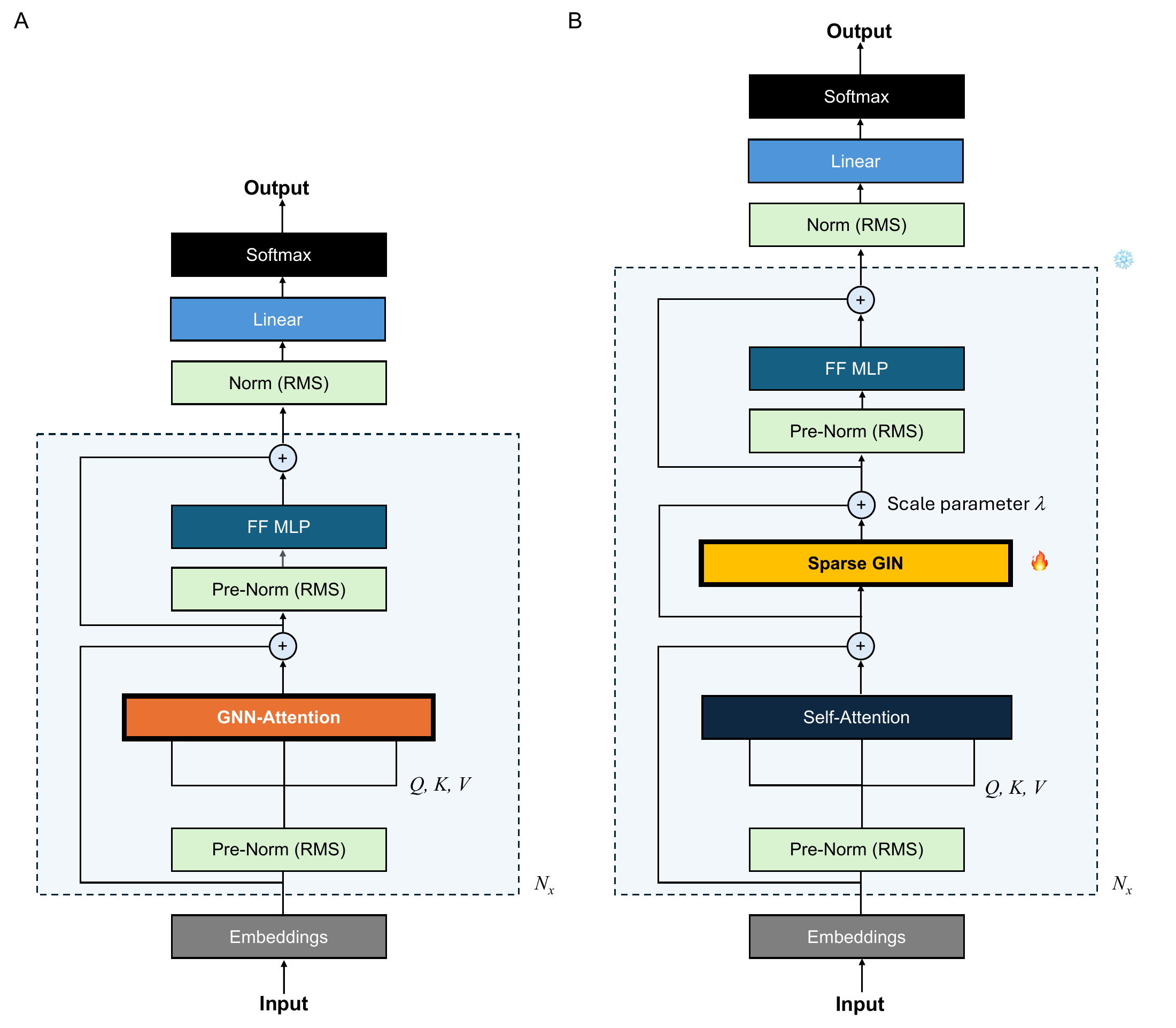}
        \caption{Decoder-only Transformer architecture (panel A), adapted here by using a GNN-based self-attention mechanism with a graph neural network (Figure~\ref{fig:GIN-Attention-flowchart} shows how GNN-Attention is constructed for the specific case of GIN-Attention). Thereby $Q$ and $K$ values are used to construct a per-head adjacency matrix, which is then used to define a causal graph. Whereas in standard Transformer models the multiplication with $V$ corresponds to a summation aggregation via a single linear layer, in GNN-Attenion we conduct more complex graph operations, including the designation of a GIN and PNA variant. As another variant (panel B) suitable for fine-tuning a pre-trained model akin to a \hlyellow{Low-Rank Adaptation (LoRA)} model~\cite{Hu2021LoRA:Models}, we introduce another option where we retain the adjacency matrix predicted by the pretrained model but instead use it to construct a sparse adjacency matrix. A Sparse GIN is defined based on this and the signal from the original attention mechanism and the GIN output is added, whereas the GIN signal is scaled by a trainable scale parameter. In this variant, the pre-trained Transformer architecture is kept intact except for the addition of the Sparse GIN block. 
     }
	\label{fig:fig_775}
\end{figure}

Decode-only Transformers~\cite{AlecRadfordImprovingPre-Training}, designed specifically for autoregressive tasks, have emerged as an efficient subclass of the Transformer architecture, focusing on multimodal tasks like text generation, image modeling, audio modeling, language modeling, and sequential data prediction, among many other modalities (Figure~\ref{fig:fig_775}). These architectures utilize an embedding layer to convert discrete tokens (which can represent diverse types of data, like text, chemical structures, images/pixels, symbols, and others) into dense vector representations, enabling the model to process flexible inputs. The core of the architecture is the self-attention mechanism, which operates causally to ensure that each token attends only to its past context, maintaining the autoregressive property essential for tasks like generative AI. Multi-head self-attention enhances the model's ability to capture diverse relationships between tokens by allowing parallel attention computations.

In addition to attention, decode-only Transformers such as those used in \texttt{LLama} foundation models~\cite{dubey2024llama3herdmodels} integrate feedforward layers (FF), usually implemented via a multi-layer perceptron (MLP), following the attention mechanism. These layers expand and transform the attention outputs, introducing non-linearity and enabling the model to learn complex patterns. The architecture also employs several residual connections and layer normalization, ensuring stability during training and facilitating the flow of gradients. Positional encodings are incorporated to inject sequence order information into the embeddings~\cite{Vaswani2017AttentionNeed,su2021roformer}, addressing the lack of inherent order in self-attention computations. Finally, the autoregressive nature of these models ensures that outputs are generated token-by-token, leveraging causal masking to prevent access to future tokens during training and inference. Together, these components create a robust and scalable framework for generative tasks.

Graph Neural Networks (GNNs)~\cite{reiser2022gnn_materials,zhang2019hierarchical,yun2019graph,velivckovic2018graph} represent another influential class of models that have effectively incorporated attention mechanisms, primarily to enhance their capacity for learning on graph-structured data. Attention in GNNs has been extensively utilized to assign varying levels of importance to nodes, edges, or subgraphs, enabling models to focus on the most relevant components for a given task~\cite{yun2019graph,velivckovic2018graph}. However, these approaches are predominantly designed for known or static graph structures, where the relationships and topology remain fixed throughout the reasoning process. These methods often lack the flexibility to dynamically adapt and reason over evolving relationships or integrate seamlessly into general-purpose Transformer architectures. Our work attempts to explore possibilities at this intersection. 

\subsection{Invoking graph theory and category theory}

\begin{figure}
	\centering
	\includegraphics[width=.8\linewidth]{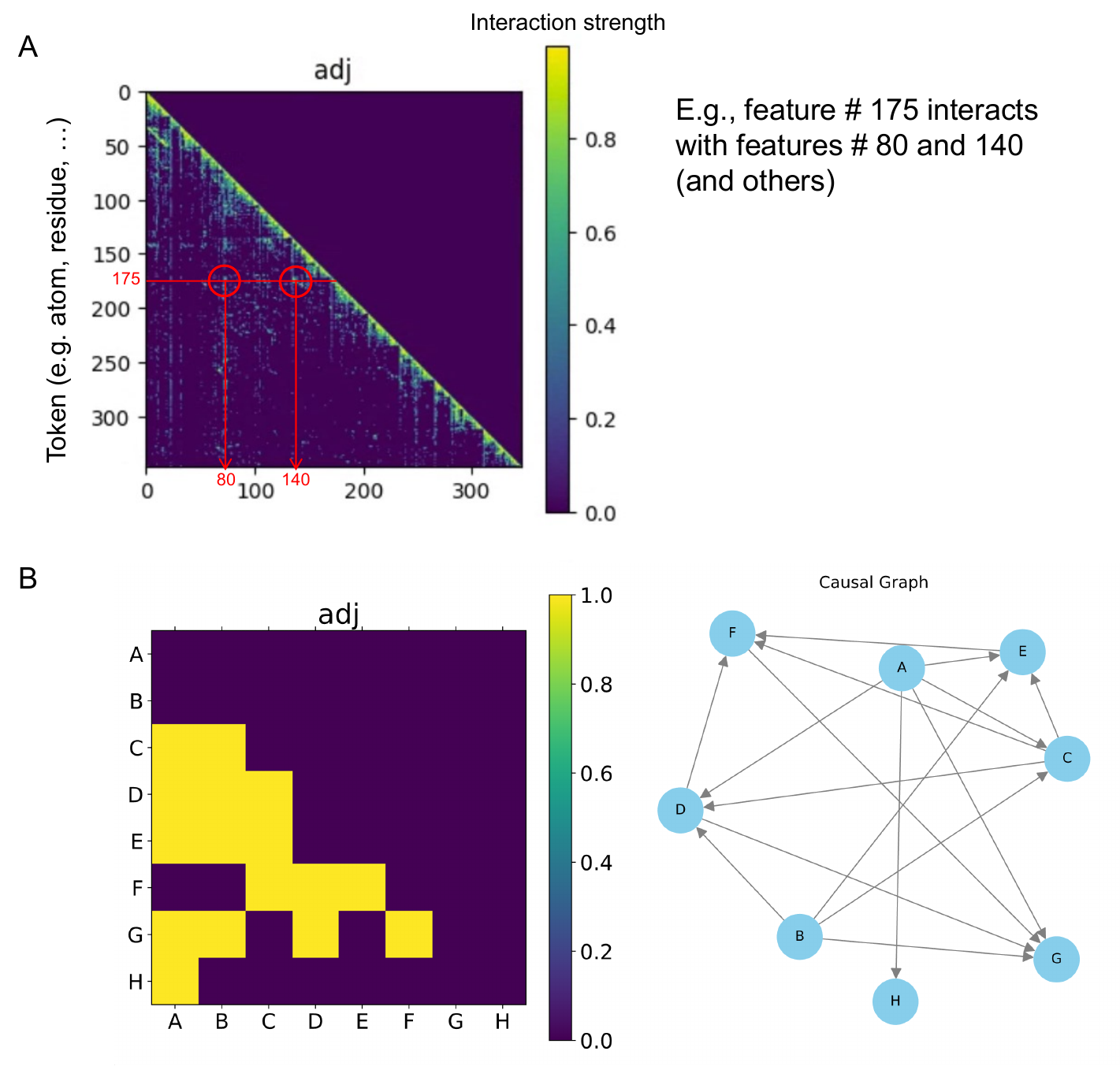}
        \caption{Visualization of adjacency matrices and interpretation of corresponding causal graphs. Panel A: Visual representation of an adjacency matrix for one specific layer and one head, extracted from a pretrained model. Panel B, left shows a large-scale adjacency matrix, where interaction strengths are color-coded, with annotations highlighting specific points of interest. Panel B, right displays the corresponding causal graph, illustrating directional relationships between nodes based on the adjacency matrix.  These visualizations provide insights into the structural and causal relationships encoded in the adjacency matrices.
     }
	\label{fig:fig_777}
\end{figure}

Category theory~\cite{Eilenberg1942GroupHomology,Eilenberg1945GeneralEquivalences}, with its abstract structures and relationships, provides a powerful lens through which we can understand the generalization capabilities of Transformers. By interpreting Transformer models as functors that map input representations to output representations, we can appreciate how these models capture abstract relationships and generalization of concepts in disparate fields. Some recent work has demonstrated universal feature activation in large models, providing evidence for such features in existing pre-trained models~\cite{cunningham2023sparseautoencodershighlyinterpretable,templeton2024scaling,Simon2024.11.14.623630,mischler2024contextual}. In particular, functors preserve structural consistency while allowing transformations across different categories, akin to how Transformers maintain contextual relevance while adapting to varied tasks and domains. This perspective sheds light on the inherent modularity of Transformers, where layers and substructures can be seen as morphisms within a category, facilitating compositional reasoning and hierarchical abstraction. Here, the attention mechanism itself can be viewed as a mapping that defines and strengthens relationships within a category, dynamically adjusting based on contextual importance, which can be interpreted as a graph as shown in Figure~\ref{fig:fig_777}. The integration of such graph-based modeling enhances this mapping by incorporating relational structures explicitly, allowing for richer representations that align with categorical transformations and hence better generalizability and predictive power, especially for scientific applications. In the context of the attention mechanism depicted in Figure~\ref{fig:fig_775}, the interpretation of attention as graph-forming operator produces a causal adjacency matrix \( A \in \mathbb{R}^{N_{\text{seq}} \times N_{\text{seq}}} \) at each layer in the model. The core graph operation in the attention mechanism is, however, linear in nature. 

The importance of isomorphic mappings as a framework for scientific generalization can be exemplified in an example of a study of biomaterials, such as spider silk, and its relation to music, drawing parallels in hierarchical structure and transformation across domains~\cite{Giesa2011ReoccurringAnalogies,Tokareva2014Structure-function-property-designSilk,Lu2023SpiderSilk,Wong2012MaterialsMusic, abbott2024flashattentionnapkindiagrammaticapproach}. Spider silk's remarkable mechanical properties arise from its hierarchical organization, spanning molecular arrangements to macroscopic fibers. In category-theoretic terms, the molecular and macroscopic structures can be viewed as objects within distinct categories, with the Transformer acting as a functor that maps molecular-level features (e.g., amino acid sequences) to macroscopic properties (e.g., tensile strength or elasticity) while preserving key relationships. Similarly, music composition involves hierarchical structures, where sequences of notes form motifs, motifs form phrases, and phrases form complete compositions. Here, a Transformer functor can map abstract patterns in one domain, such as the hierarchical structures of biomaterials, to analogous patterns in another, such as the rhythmic or melodic structures in music. The attention mechanism, by dynamically constructing adjacency matrices, captures critical intra- and inter-level relationship -- whether between molecular components in silk or between notes in a melody. Integrating graph-based neural networks into this framework enhances the mapping by explicitly propagating relational structures, enabling the discovery of universal principles underlying both biomaterials and music. This perspective highlights the power of category theory in unifying seemingly disparate domains through abstract structural consistency and transformation.

In this interpretation, graph theory bridges these categorical insights into practical implementations. Attention mechanisms can be seen as dynamically evolving adjacency matrices, where learned relationships between input elements correspond to edge weights in a graph (Figure~\ref{fig:fig_777}). Once the adjacency matrix \( A \) is constructed, it is used to perform a simple weighted sum over the value matrix \( V \in \mathbb{R}^{N_{\text{seq}} \times d_v} \), where each row of \( V \) represents the value vector corresponding to a token. Leveraging this insight, one can hypothesize that enhancing these adjacency matrices with powerful Graph Neural Network (GNN), such as Graph Isomorphism Networks (GIN)~\cite{xu2019how} or Principal Neighborhood Aggregation (PNA) models~\cite{corso2020principal}, can result in improved learning, abstraction and generalization capabilities. By strengthening the ability to capture graph isomorphism properties, GNNs can align relational abstractions within the Transformer framework, further solidifying contextual relationships through the lens of isomorphic relational discoveries. This approach hypothesizes that enhancing attention's capacity to capture and manipulate structural relationships will enable models to generalize better across domains and tasks. That is, unlike traditional graph-attention methods, which typically focus on local neighborhoods or hierarchical scales (and known graph structures), our approach integrates a fine-tuning strategy using sparse graph neural networks. By interpreting dynamically generated attention matrices as adjacency graphs and enabling their dynamic adjustment during training, we achieve a versatile framework that bridges sequential and relational data. Furthermore, our theoretical framing of Transformers as implicitly graph-based models offers novel insights into their generalization capabilities, paving the way for more interpretable and adaptable architectures that extend beyond the constraints of existing graph-attention models.

This perspective not only deepens our theoretical understanding of Transformers but also suggests pathways for structural innovation in how this class of neural networks can be adapted for scientific applications. For example, leveraging categorical constructs such as natural transformations could inform the design of more robust attention mechanisms, enabling seamless adaptation across increasingly complex data landscapes. The core idea behind this paper is to use concepts inspired by category theory with a practical implementation to synthesize an improved framework to train and fine-tune Transformer models. Within this scope, this work focuses on architectural details tested on a variety of sample training sets, incorporating graph-based relational structures to enhance the generalization and abstraction capabilities of the Transformer architecture.

\subsection{Outline}

We explore the theoretical foundations of attention mechanisms and propose a series of alterations to the conventional architecture. We interpret the Transformer's attention mechanism as a graph generating operator and enhance its structure by introducing more complex graph neural networks such as the Graph Isomorphic Neural network (GIN).  Additionally, we develop a novel fine-tuning technique that can serve as a replacement or complement to LoRA with graph neural networks, interpreting the adjacency matrix predicted by conventional attention as an input to a graph neural network. This approach enhances the model's ability to learn from both sequential data and structured graph data. Through these enhancements, our aim is to provide a robust framework for leveraging graph structures in Transformer-based models, offering valuable insights into enhancing attention mechanisms.

\section{Results and Discussion}

\subsection{Theoretical foundations: Graph-aware attention}

We first review details of the conventional attention mechanism and then describe our revised approach to introduce graph-aware or GNN-Attention in various flavors. 

\subsubsection{Interpreting Attention as a linear GNN}

The attention mechanism can be interpreted as a form of message passing on a fully connected graph, where each token in the input sequence corresponds to a node, and the learned attention scores define the weighted edges. Mathematically, the adjacency matrix \( A \in \mathbb{R}^{N_{\text{seq}} \times N_{\text{seq}}} \) is computed using scaled dot-product attention:

\begin{equation}
A_{ij} = \text{softmax}\left(\frac{Q_i \cdot K_j^\top}{\sqrt{d_k}}\right),
\end{equation}
where \( Q, K \in \mathbb{R}^{N_{\text{seq}} \times d_k} \) are the query and key matrices, derived from the input embeddings \( X \) as \( Q = X W_Q \) and \( K = X W_K \), with \( W_Q, W_K \in \mathbb{R}^{d_{\text{model}} \times d_k} \) being learnable projection matrices. The softmax ensures that the rows of \( A \) sum to 1, effectively normalizing the attention scores into a probability distribution. The value matrix \( V \in \mathbb{R}^{N_{\text{seq}} \times d_v} \) is then aggregated using \( A \):

\[
Y = A V,
\]

where \( Y \in \mathbb{R}^{N_{\text{seq}} \times d_v} \) represents the output of the attention layer. This aggregation via $V$ can be viewed as a graph neural network operation, where \( A_{ij} \) serves as the weight of the edge from node \( i \) to node \( j \), and \( V_j \) is the feature vector of node \( j \). The output \( Y_i \) for each node \( i \) is a weighted sum of its neighbors' features, aligned with the aggregation strategy used in graph neural networks.

In the context of multi-head attention, multiple attention heads independently compute adjacency matrices \( A_h \) and corresponding outputs \( Y_h = A_h V_h \), where \( h \in \{1, \dots, H\} \) denotes the attention head and \( V_h = X W_{V_h} \).  That is, each attention head computes an independent adjacency matrix \( A_h \) and performs GIN aggregation. The resulting embeddings are concatenated across heads and linearly transformed to produce the final output, allowing each head to capture diverse relational patterns while maintaining computational efficiency.

The outputs from all heads are then concatenated and linearly transformed over all heads:

\[
Y_{\text{multi-head}} = \left(\bigoplus_{i=1}^H Y_i \right ) W_O,
\]

with \( W_O \in \mathbb{R}^{(H \cdot d_v) \times d_{\text{model}}} \) as the output projection matrix. The per-head mechanism allows the model to attend to different representations of the input data simultaneously. Specifically, this interpretation of attention as a graph operation highlights how the learned adjacency matrix \( A \) dynamically adjusts to capture task-specific relationships. By incorporating multi-head attention, the mechanism effectively performs diverse message-passing computations, enriching the representation of node (token) features.

\subsubsection{Expressive graph attention through a Graph Isomorphism Network (GIN): GIN-Attention}

We now build on this concept and propose an enhanced attention mechanism that expands the traditional scaled dot-product attention by replacing the linear aggregation of value vectors \( V \) with a Graph Isomorphism Network (GIN)-based process. Additionally, this mechanism introduces a sharpening parameter to the attention computation and allows for the use of alternative activation functions beyond Softmax. The detailed steps of the algorithm are as follows.

In its most basic form, the adjacency matrix \( A \) is computed using a modified Softmax function, which incorporates a learnable sharpening parameter \( \alpha \) (we can view the sharpening parameter as a dial on a microscope to focus in on key relationships):

\begin{equation}
A_{ij} = \text{Softmax}\left( \frac{\alpha \cdot Q_i \cdot K_j^\top}{\sqrt{d_k}} \right),
\end{equation}
where:
\[
Q = X W_Q, \quad K = X W_K,
\]
and \( W_Q, W_K \in \mathbb{R}^{d_{\text{model}} \times d_k} \) are learnable projection matrices. The sharpening parameter \( \alpha \) dynamically adjusts the focus of the attention distribution. A higher \( \alpha \) leads to sharper attention, while a lower \( \alpha \) produces a smoother distribution. Alternatively, the Softmax operation can be replaced with other activations, such as element-wise sigmoids:
\[
\text{Sigmoid: } A_{ij} = \sigma\left( \beta \cdot (Q_i \cdot K_j^\top - \tau) \right),
\]
or the Sharp Softplus function:
\[
\text{Sharp Softplus: } A_{ij} = \frac{\log\left(1 + e^{\beta \cdot (Q_i \cdot K_j^\top - \tau)}\right)}{\beta},
\]
where, \( \beta \) controls the sharpness, and \( \tau \) represents a learnable threshold.

The adjacency matrix \( A \) can be further refined based on various modes, such as applying thresholding, sparsification (e.g., top-\( k \)), or additional activations. These transformations adapt \( A \) to better represent the graph structure before aggregation. For instance:
\[
A' = \text{ReLU}(A - \tau), \quad A' = \text{top-}k(A, k).
\]

Causality is enforced by applying a lower triangular mask to ensure that each node only attends to itself and its predecessors.

Once the adjacency matrix \( A' \) is computed, the GIN aggregation is applied. The updated node embeddings \( X' \) are computed as:
\[
X' = \text{MLP}\left( \epsilon \cdot X + A' X \right),
\]
where \( \epsilon \) is a learnable parameter that scales the contribution of the current node embeddings \( X \), and \( A' X \) aggregates information from the neighboring nodes. 

The original GIN formulation~\cite{xu2019how} updates node embeddings as:

\[
X' = \text{MLP}\left((1 + \epsilon) \cdot X + A X\right),
\]

where \( 1 \cdot X \) explicitly includes the self-loop contribution, and \( \epsilon \) is a learnable parameter controlling its importance.
 Unlike traditional GIN formulations, the term \( 1 \cdot X \) is omitted here because the skip connection around the attention block  inherently captures this self-loop (see, e.g. Figure~\ref{fig:fig_775}). The MLP consists of two linear transformations, a non-linear activation (e.g., SiLU), and optional dropout and normalization layers (details below).

In the multi-head setting used here, each head \( h \) computes its own adjacency matrix \( A_h' \) and performs GIN-based aggregation independently:
\[
X'_h = \text{MLP}_h\left( \epsilon_h \cdot X_h + A_h' X_h \right).
\]
The outputs from all heads are concatenated and ultimately, linearly transformed:
\[
Y= \text{Concat}(X'_1, X'_2, \dots, X'_H) W_O,
\]
where \( W_O \in \mathbb{R}^{(H \cdot d_k) \times d_{\text{model}}} \) is an optional learnable output projection matrix (if not used, the aggregated output is used without further transformation). This design allows each head to focus on different relational patterns within the input, enhancing the overall expressiveness of the attention mechanism.

The GIN-aggregated outputs \( X' \) are combined with the residual connection and normalized using RMSNorm.

\begin{figure}
\sffamily 
\scriptsize
\sansmath 

\begin{tikzpicture}[
    every node/.style={font=\sffamily, align=center},
    block/.style={draw, rectangle, rounded corners, minimum height=1.2cm, minimum width=4.5cm, align=center, font=\small},
    attention/.style={draw, rectangle, fill=salmon!50, rounded corners, minimum height=1.2cm, minimum width=4.5cm, align=center, font=\small},
    gnn/.style={draw, rectangle, fill=lightblue!50, rounded corners, minimum height=1.2cm, minimum width=4.5cm, align=center, font=\small},
    arrow/.style={thick,->,>=stealth},
    dashedarrow/.style={thick,->,>=stealth, dashed}
]

\node[block, fill=lightgray] (input) at (0, 0) {Input Hidden States \\ $X$};
\node[attention] (attention) at (0, 2) {Self-Attention  $\frac{Q_i \cdot K_j^\top}{\sqrt{d_k}}$};
\node[block, fill=yellow!50] (adj_process) at (0, 4) {Adjacency Matrix \\ Processing \\ $A'$};
\node[gnn] (gin) at (0, 6) {GIN Layer \\ $X' = \epsilon \cdot X + A'X$};
\node[block, fill=lightgreen!50] (mlp) at (0, 8) {Multi-layer Perceptron (MLP) \\ $Y = \text{MLP}(X')$};
\node[block, fill=gray!20] (output) at (0, 10) {Output Hidden States \\$Y$};

\draw[arrow] (input) -- (attention) node[midway, right] {Query, Key};
\draw[arrow] (attention) -- (adj_process) node[midway, right] {Adjacency Matrix $A$ (per head)};
\draw[arrow] (adj_process) -- (gin) node[midway, right] {Refined $A'$};
\draw[arrow] (gin) -- (mlp) node[midway, right] {GIN Output $X'$};
\draw[arrow] (mlp) -- (output);

\node[right=2cm of adj_process, font=\small, text width=4cm, align=left] {Options: Sharpening, \\ Softmax, \\ Thresholding, \\ Sigmoid, \\ Sharp Softplus};
\node[right=2cm of gin, font=\small, text width=4cm, align=left] {Sum Aggregation};
\node[right=2cm of mlp, font=\small, text width=4cm, align=left] {MLP for nonlinear expressiveness};

\end{tikzpicture}

\caption{Construction of the GIN-Attention mechanism. The flowchart shows how input embeddings in the hidden states in each layer in the  transformer via self-attention are used to construct the attention matrix. The output is processed further before aggregation and GIN-MLP application. The alternative PNA processing discussed in the paper is done in a conceptually similar way, except that we use query, key and value projections followed by developing up to four distinct \hlyellow{aggregations} that are concatenated and then projected back into the hidden dimension via a MLP.}
\label{fig:GIN-Attention-flowchart}

\end{figure}
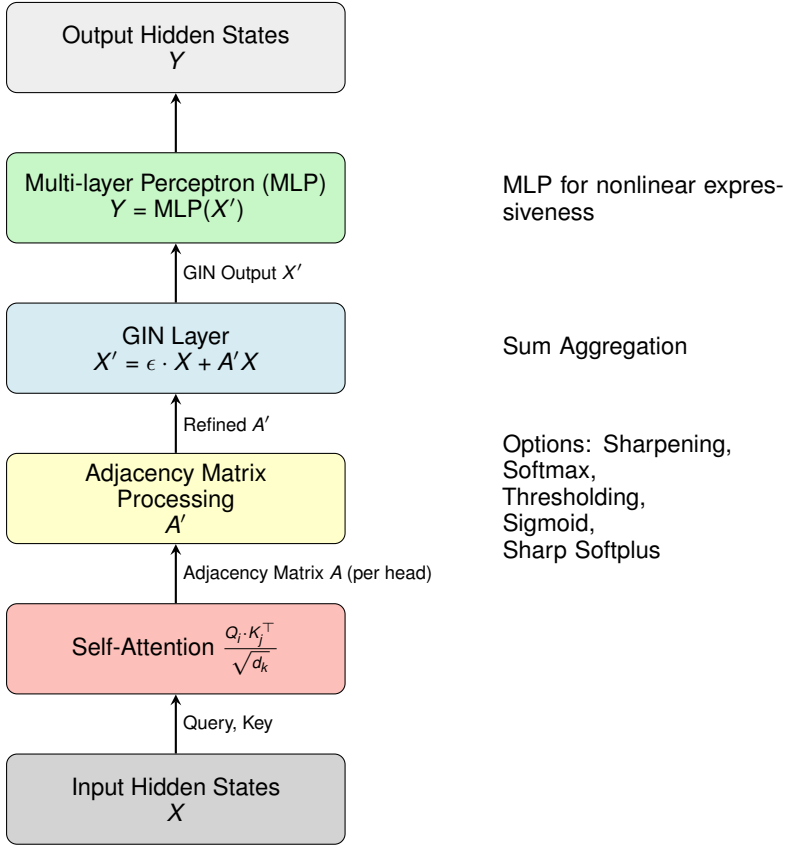

The overall framework is visualized schematically in Figure~\ref{fig:GIN-Attention-flowchart}. It systematically integrates graph-based modeling into the attention mechanism, enabling the Transformer to capture richer relational structures and achieve better generalization.

\paragraph{Details on GIN layer design}

The Graph Isomorphism Network (GIN) in this work is constructed to enhance relational modeling capabilities. The GIN aggregates node features \( X \) by applying the adjacency matrix \( A' \) (defined for each head, but this detail \hlyellow{is omitted here} since we introduce the general GIN layer architecture) to perform a summation aggregation  from neighboring nodes and updates the embeddings as:

\[
X' = \text{MLP}(\epsilon \cdot X + A' X),
\]

where \( \epsilon \) is a learnable parameter controlling the self-loop contribution. The MLP is designed with two linear layers and intermediate transformations, implemented as:

\[
\text{MLP}(z) = \text{Linear}_2(\text{Dropout}(\text{SiLU}(\text{Norm}(\text{Linear}_1(z))))),
\]

where:
\[
\text{Linear}_1: \mathbb{R}^{d_{\text{in}}} \to \mathbb{R}^{d_h}, \quad \text{Linear}_2: \mathbb{R}^{d_h} \to \mathbb{R}^{d_{\text{in}}},
\]
and \( d_h = \gamma \cdot d_{\text{in}} \), with \( \gamma \) being a configurable multiplier that determines the intermediate hidden dimension of the MLP layer. This multiplier allows for control over the capacity of the MLP, enabling adjustments to the model's expressiveness based on task complexity. \( \text{Norm} \) refers to RMSNorm, which stabilizes the feature representations, while SiLU provides non-linearity. Dropout is applied after activation to regularize the model and prevent overfitting.

We note that one could, in principle, feed an external adjacency matrix to be added to the values identified by the attention mechanisms. While this is not explored here, it could be subject of further research, along with trainable methods to construct adjacency matrix that could replace or complement the attention mechanism focused on in this work.

\subsubsection{Extension to Principal Neighborhood Aggregation (PNA)-Attention}

The attention mechanism can be extended by replacing the conventional value aggregation with more complex neighborhood aggregation strategies, inspired by Principal Neighborhood Aggregation (PNA)~\cite{corso2020principal}. This approach introduces multiple aggregators—such as sum, mean, max, and variance—to capture diverse relational patterns within the input graph structure.

As before, the adjacency matrix \( A \) is first computed using scaled dot-product attention, with the option to refine it via thresholding, sparsification, or alternative activation functions. Once the adjacency matrix \( A \) is processed, the aggregators compute the characteristics of the neighborhood of each node. For example:
\[
\text{Sum Aggregator: } \text{sum\_agg} = A X,
\]
\[
\text{Mean Aggregator: } \text{mean\_agg} = \frac{A X}{\text{deg}},
\]
where \( \text{deg} = A \mathbf{1} \) is the degree of each node, and \( \mathbf{1} \) is the all-ones vector. Additional aggregators include:
\[
\text{Max Aggregator: } \text{max\_agg}_i = \max_{j \in \text{neighbors}(i)} X_j,
\]
\[
\text{Variance Aggregator: } \text{var\_agg} = \text{mean\_of\_squares} - (\text{mean\_agg})^2.
\]

The outputs of these aggregators are concatenated for each node to form a feature vector that combines information across multiple aggregation modes:
\[
\text{combined} = \text{Concat}(\text{sum\_agg}, \text{mean\_agg}, \text{max\_agg}, \text{var\_agg}).
\]

This combined feature vector is then passed through a Multi-Layer Perceptron (MLP) to produce the updated node embeddings. The MLP is defined as:
\[
\text{MLP}(z) = \text{Linear}_2(\text{SiLU}(\text{Linear}_1(z))),
\]
where \( \text{Linear}_1 \) projects the combined features to a hidden dimension, and \( \text{Linear}_2 \) maps it back to the original dimension.

As in the GIN-Attention case, in the multi-head setting, each head \( h \) computes its own adjacency matrix \( A_h \) and applies the aggregation independently. The outputs from all heads are concatenated and linearly transformed:
\[
Y = \text{Concat}(X'_1, X'_2, \dots, X'_H) W_O,
\]
where \( W_O \) is the learnable output projection matrix. This approach enables each head to learn distinct relational patterns, enriching the overall representation.

To ensure causality, a lower triangular mask is applied to \( A \), ensuring that each node only attends to itself and its predecessors. Additionally, the mechanism allows for blending the original attention output with the aggregated output using a learnable residual blending ratio:
\[
Y = (1 - \lambda) \cdot Y_{\text{original}} + \lambda \cdot Y_{\text{aggregated}},
\]
where \( \lambda \in [0, 1] \) is a learnable blending parameter.

This extended framework systematically integrates multiple neighborhood aggregation strategies, enhancing the flexibility and expressiveness of the attention mechanism while maintaining computational efficiency.

As before, the adjacency matrix \( A \) can undergo flexible processing to adapt to the desired relational structure. If conventional softmax normalization is not applied, alternatives such as ReLU can be used directly on the attention scores, enabling sparsity while preserving non-negative weights. Further, the algorithm allows for advanced transformations, including thresholding, sharp softplus activation, sigmoid-based scaling, or top-\( k \) sparsification.  These options allow that the adjacency matrix can dynamically adapt to various tasks while retaining causality through the application of a lower triangular mask.

As before we could in principle  feed an external adjacency matrix to be added to the values identified by the attention mechanisms.

\subsection{Experimental results: Graph-aware attention}

We present several case studies to test the performance of GIN-Attention and PNA-Attention versus standard linear attention. For all tests we use a consistent model size of ~25 M parameter, the same training data, with the same training hyperparameters. 

\begin{figure}
	\centering
	\includegraphics[width=.8\linewidth]{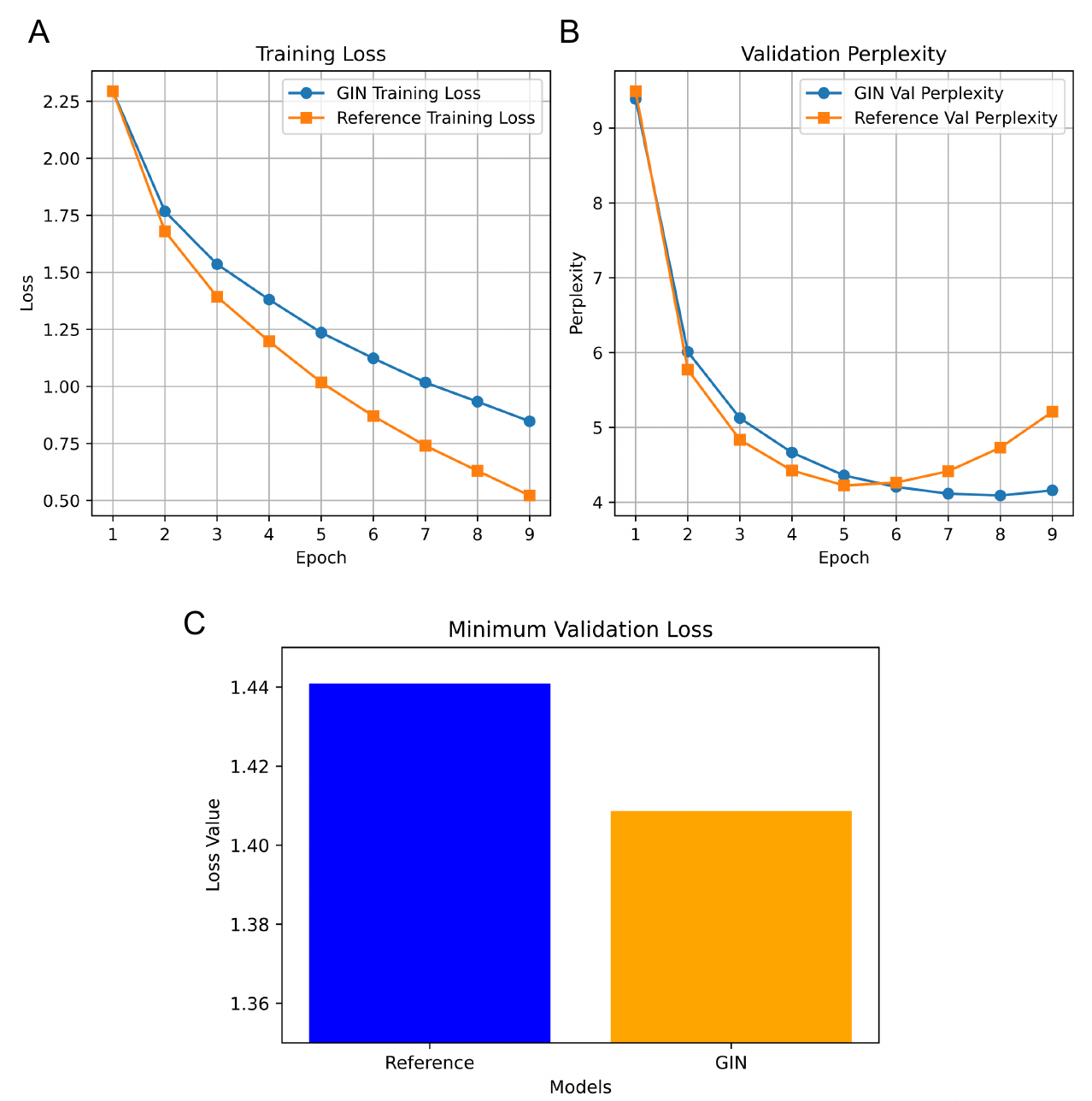}
        \caption{Training and validation performance of the regular transformer model (identified as ``Reference'' model)) and the GIN model. A, Training loss comparing the regular transformer and GIN model, over training epochs. B, Validation perplexity comparing the regular transformer and GIN model, over training epochs. C, Minimum validation loss measured across all epochs. The minimum validation loss is found in epoch 5 for the regular transformer model, and in epoch 8 for the GIN model. }
	\label{fig:fig_101}
\end{figure}

Figure~\ref{fig:fig_101} shows the performance of the regular transformer model, identified as ``Reference'' model) and the GIN-Attention model. The figure shows training loss comparing the regular transformer and GIN model, over training epochs along with validation perplexity comparing the regular transformer and GIN model, over training epochs. It can be seen that the GIN-Attention model clearly outperforms the standard attention model. Both, the minimum validation loss is found to be lower for GIN-Attention and the model shows significantly less overfitting. The lowest validation loss is found in epoch 5 for the regular transformer model, and in epoch 8 for the GIN model. 

\begin{figure}
	\centering
	\includegraphics[width=0.65\linewidth]{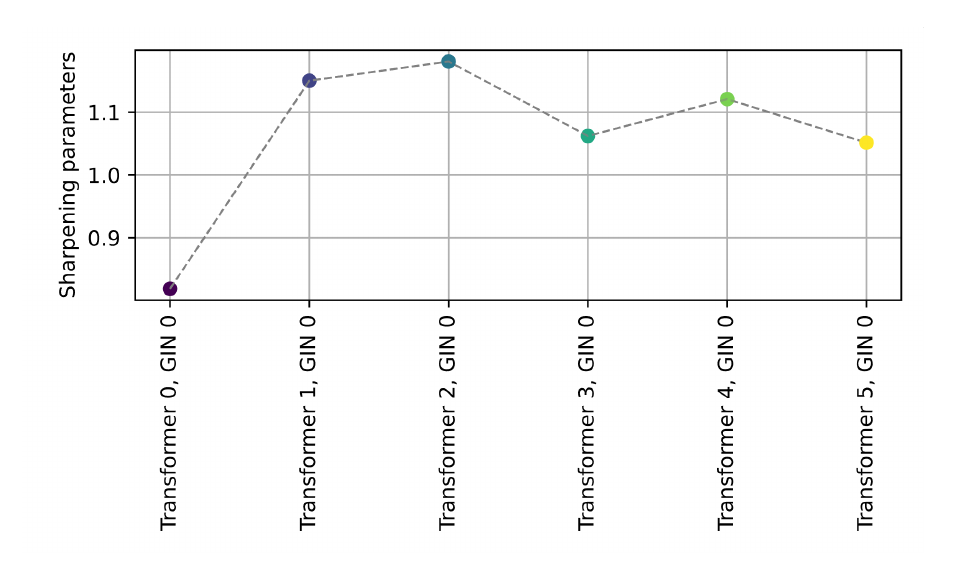}
        \caption{
        The distribution of the sharpening parameter \( \alpha_i \) across all layers \( i \) in the GIN model at the end of training. The sharpening parameter \( \alpha_i \) controls the focus of the attention mechanism by scaling the logits before applying the softmax function. A value of \( \alpha_i = 1.0 \) corresponds to the standard softmax behavior, where no additional sharpening or smoothing is applied. The variation of \( \alpha_i \) indicates how different layers adjust their focus during training. Layers with \( \alpha_i > 1.0 \) exhibit sharper attention distributions, focusing more strongly on specific tokens, while layers with \( \alpha_i < 1.0 \) produce smoother attention distributions, allowing a more even consideration of all tokens. This behavior reflects the adaptive nature of the GIN model in optimizing attention mechanisms for different layers to improve overall performance. All models are constructed to have approximately the same number of parameters, 25M.
    }
	\label{fig:fig_102}
\end{figure}

Among the graph-aware attention Transformer models, the best performance was found for GIN-Attention with Softmax activation and the use of a trainable per-layer sharpening parameter  \( \alpha_i \). The final distribution of the sharpening parameter across the transformer layers is shown in  Figure~\ref{fig:fig_102}, revealing an interesting distribution of how sharpening is used strategically distributed across the model layers.

\begin{figure}
	\centering
	\includegraphics[width=1.\linewidth]{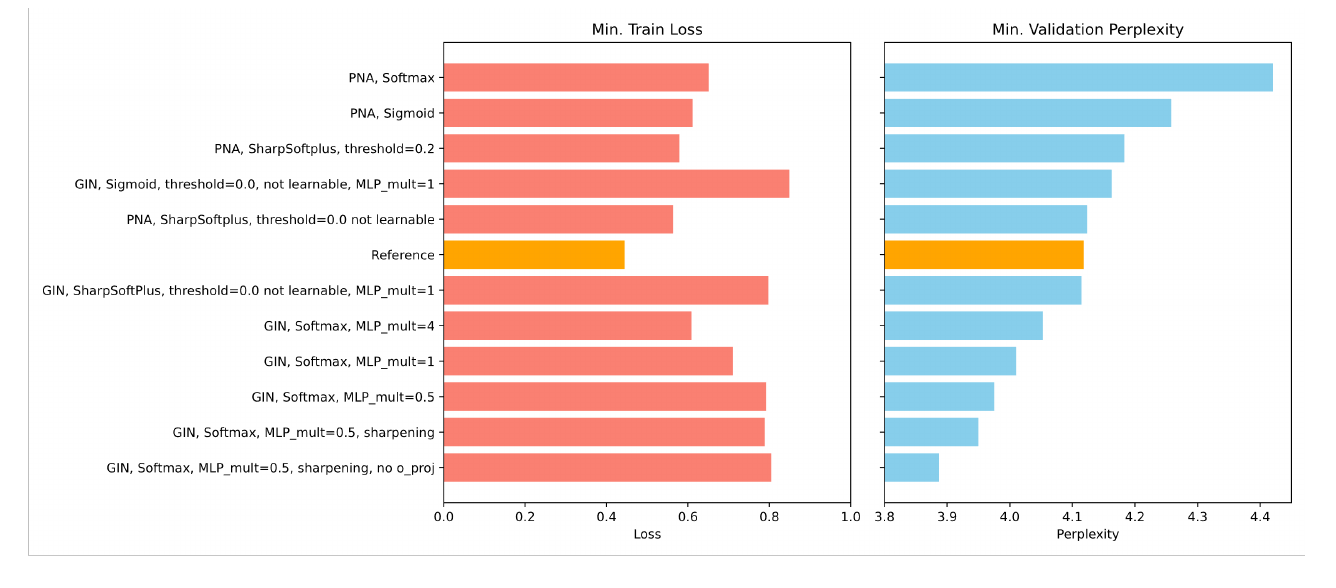}
        \caption{Minimum training loss and minimum validation perplexity measured, across a variety of cases. The case identified as ``Reference'' is a regular transformer architecture. Cases considered include PNA attention, GIN attention, and variations within each scenario. The best performing model with lowest validation perplexity is GIN attention with softmax and a MLP multiplier \( \gamma \)  of 0.5, with trainable sharpening parameter. Except for one case, all GIN model architectures perform better than the reference standard attention. None of the PNA architectures improves upon the reference case, suggesting that this architectural concept is not viable.}
	\label{fig:fig_120}
\end{figure}

Figure~\ref{fig:fig_120} shows the minimum training loss and minimum validation perplexity across a variety of experiments for a systematic comparison. As above, the case identified as ``Reference'' is a regular transformer architecture, as before. In addition to GIN-Attention, cases considered include PNA-Attention, and variations within each model architecture. We find that the best performing model with lowest validation perplexity is GIN attention with Softmax and a MLP multiplier \( \gamma \) of 0.5, with trainable sharpening parameter and no \texttt{o\_proj} layer.

We also find that except for one case, all GIN model architectures perform better than the reference standard attention. None of the PNA architectures improves upon the reference case, suggesting that this architectural concept is not viable (at least for the cases explored). We present the results here for comparison and to leave this direction of research for further investigation.

\begin{figure}
	\centering
	\includegraphics[width=0.7\linewidth]{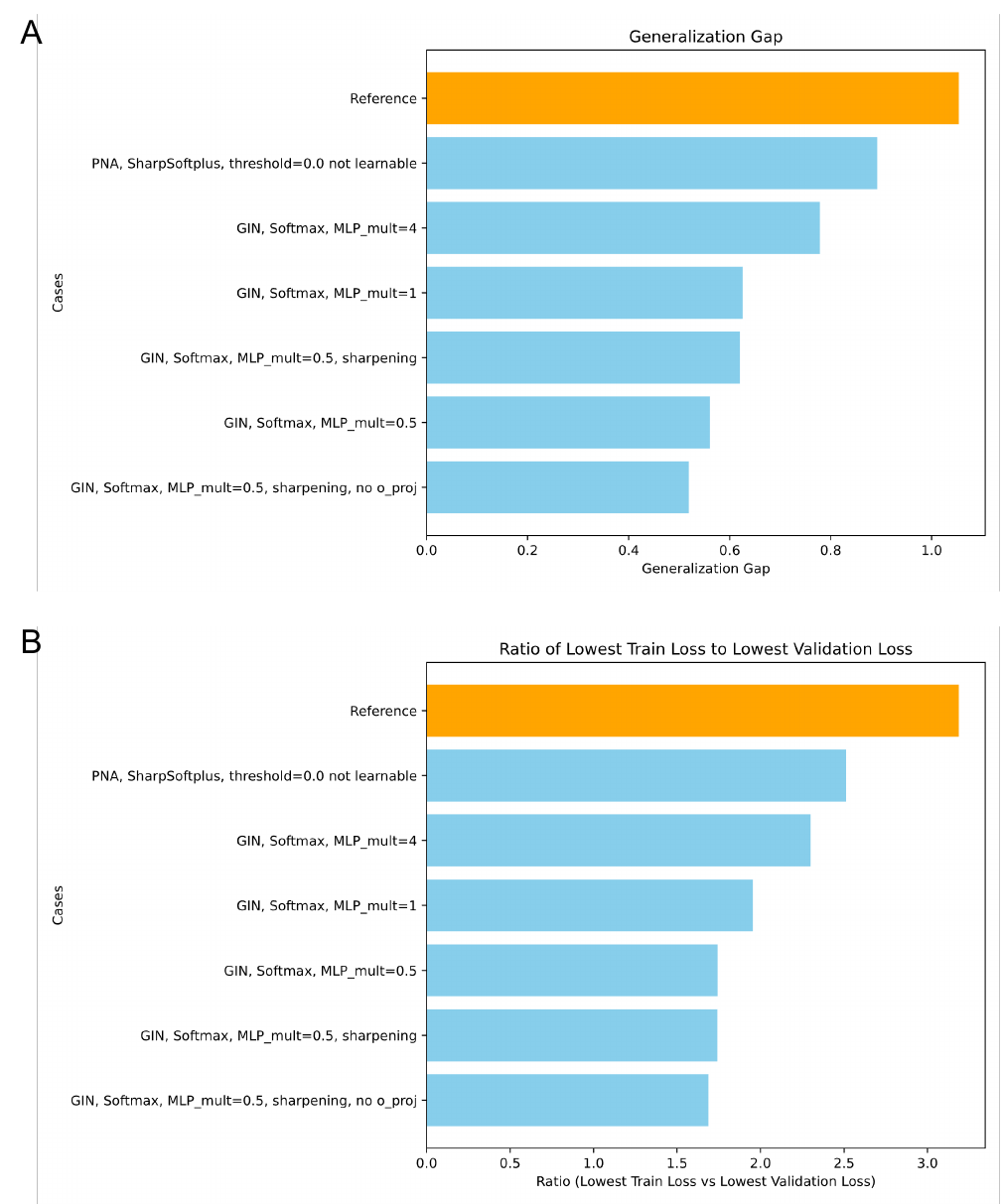}
        \caption{Further training dynamics analysis. Panel A: Generalization gap for selected cases that perform well overall, measured after 9 training epochs. The reference model shows the highest generalization gap, indicating overfitting. Models using GIN (Graph Isomorphism Network) with Softmax and varying MLP multipliers demonstrate reduced generalization gaps, with the GIN configuration using a multiplier \( \gamma \) of 0.5 and sharpening achieving without o\_proj one of the lowest gaps. The PNA configuration with SharpSoftplus activation and a fixed threshold also exhibits improved generalization compared to the reference. This comparison highlights the effect of architectural choices on model generalization. Panel B: Ratio of lowest training loss to lowest validation loss achieved. The lowest ratio is also found for the GIN model using a multiplier \( \gamma \) of 0.5 and sharpening achieving without o\_proj.}
	\label{fig:fig_121}
\end{figure}

Next, Figure~\ref{fig:fig_121} depicts the generalization gap and the ratio of the lowest training loss to the lowest validation loss for select cases that perform well overall. We find that the reference model shows the highest generalization gap, indicating overfitting. Models using GIN-Attention with Softmax and varying MLP multipliers demonstrate reduced generalization gaps, with the GIN configuration using a multiplier \( \gamma \) of 0.5 and sharpening without \texttt{o\_proj}  achieving the best performance consistent with earlier results. We note that GIN-Attention with SharpSoftplus activation and a fixed threshold also exhibits improved generalization compared to the reference, but falls short compared to the other cases. This comparison highlights the effect of architectural choices on model generalization and gives us insights into which specific approach works best.

Looking at the results, we observe an important effect of the GIN MLP multiplier ratio \( \gamma \). To further analyze this behavior systematically, Figure~\ref{fig:fig_121} shows the minimum validation perplexity as a function of \( \gamma \). We find that the relationship between \( \gamma \) and validation perplexity for various configurations follow a trend, indicating an increase in validation perplexity as the GIN MLP multiplier ratio grows. Configurations with lower MLP ratios (specifically: \( \text{MLP\_mult=0.5} \)) exhibit better validation perplexity, suggesting a trade-off between multiplier ratio and generalization. We suspect that this choice has a regularization effect on the training performance. 

\begin{figure}
	\centering
	\includegraphics[width=0.7\linewidth]{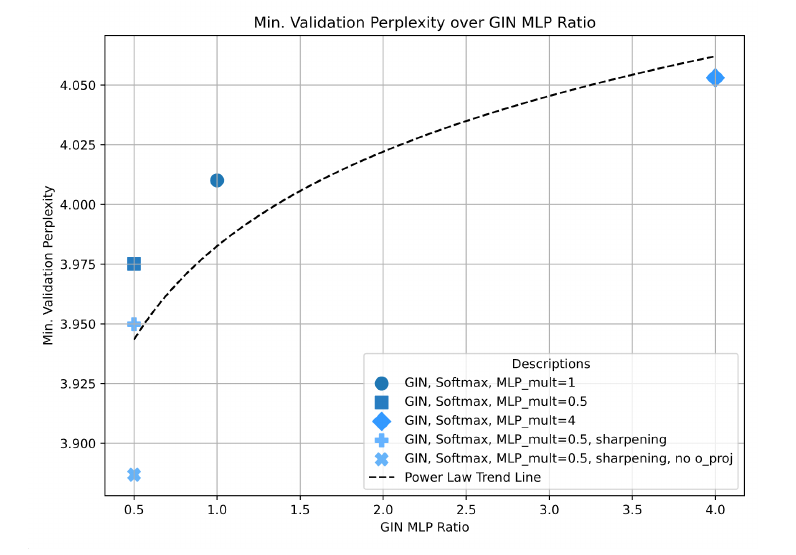}
        \caption{Minimum validation perplexity as a function of the GIN MLP multiplier ratio \( \gamma \). The plot demonstrates the relationship between the GIN MLP multiplied \( \gamma \)  and validation perplexity for various configurations: \( \text{MLP\_mult=0.5, sharpening} \), \( \text{MLP\_mult=0.5} \), \( \text{MLP\_mult=1} \), and \( \text{MLP\_mult=4} \). The data points are fitted with a power law trend line, indicating an increase in validation perplexity as the GIN MLP multiplier ratio grows. Configurations with lower MLP ratios (e.g., \( \text{MLP\_mult=0.5} \)) exhibit better validation perplexity, suggesting a trade-off between multiplier ratio and generalization.}
	\label{fig:fig_122}
\end{figure}

The results demonstrate a clear advantage of using GIN-Attention as a strategy to revise the conventional attention mechanisms in pretraining strategies, here demonstrated for protein modeling tasks. 

\subsection{Sparse graph-aware attention as a fine-tuning strategy}
\label{sparse-gin-formalism}

The earlier discussion showed how expanding linear attention to include more expressive graph operations such as GIN-Attention can increase performance. However, training Transformer models from scratch can be a time-consuming and expensive process. Here we explore how the general concept can be used to develop a fine-tuning strategy that allows us to use a pre-trained transformer model and adapt it to a particular set of training data. 

In this method, a Graph Neural Network (GNN) is integrated into the Transformer architecture immediately after the standard self-attention mechanism, enhancing the model's ability to capture relational structures within the data, but geared towards a fine-tuning approach similar to \hlyellow{Low-Rank Adaptation (LoRA)} that adds shallow layers to a model~\cite{Hu2021LoRA:Models,Buehler2024X-LoRA:Design,mao2024survey_lora_llms}. The adjacency matrix \( A \) is derived from the attention weights computed during \hlyellow{the} standard self-attention step. However, it is refined based on specific configurations, such as thresholding to achieve sparsification to yield efficient computational strategies. 

We experimented with a variety of aggregation strategies and found summation mechanisms to work well overall, while being easy to implement. First, we sum all per-head adjacency matrices $A_h$ while clamping values at one, providing an aggregated view of all relationships identified by the attention mechanism. The effective aggregated adjacency matrix \( A \) is computed by summing the per-head adjacency matrices \( A_h \) and applying a clamping function to ensure the values remain within a specified range:
\begin{equation}
A_{\text{agg}} = \min\left(1, \sum_{h=1}^{H} A_h \right),
\end{equation}
transforming $H$ individual adjacency matrices to a single, \hlyellow{aggregated} adjacency matrix $A$. A sharpening function is applied:
\[
A' = \textrm{sigmoid}\left(\alpha(A_{\text{agg}}  - \tau)\right),
\]
where \( \alpha \) controls the sharpness of the sigmoid and \( \tau \) is a threshold that allows for further fine-tuning of the mapping function.

Afterwards, discrete thresholding is applied:
\[
A' =
\begin{cases} 
A'_{ij}, & \text{if } A'_{ij}  > \epsilon \\
0, & \text{otherwise}
\end{cases}
\]
Alternatively, other sparsification strategies like top-\( k \) selection may be used. To ensure causality, a lower triangular mask is applied, restricting each node to attend only to itself and its predecessors.

This approach systematically combines the expressiveness of the GNN with the contextual representations learned by the attention mechanism.

The refined adjacency matrix \( A' \) is passed along with the transformer hidden states \( X \) into the \hlyellow{GNN layer for further processing}. The updated embeddings are computed as:
\[
X' = X + \text{GNN}(X, A') \cdot \lambda,
\]
where \( \lambda \) is a learnable scaling parameter that modulates the contribution of the GNN output to the final embeddings. This scale ensures the GNN's impact is dynamically adjusted during training, allowing for effective blending of relational features with the transformer hidden states. We typically chose initial values for  \( \lambda \) as a linear curve over model layers from 0 to a finite value, e.g. 0.5 in the last layer. 

After the GNN aggregation, the output \( X' \) is combined with the residual from the input hidden states. Additionally, a normalization step is applied to ensure numerical stability and consistent scaling of the features:
\[
X'' = \text{Norm}(X'),
\]
where \( \text{Norm} \) refers to RMSNorm or another normalization technique.

The final output of the layer is the normalized, GNN-augmented hidden states:
\[
Y = X''.
\]

By integrating the GIN immediately after the self-attention step, this approach refines the relational representations learned by the attention mechanism before the signal is fed to the feed-forward (FF) MLP block. The explicit inclusion of \( \lambda \) ensures that the GNN's impact can be adaptively controlled, providing a robust and flexible mechanism for capturing complex dependencies within the input data.

\paragraph{GIN Layer Implementation}

The Sparse GIN framework implements a Graph Isomorphism Network layer to refine node representations using causal graph structures. Consistent with the \hlyellow{reference} approach we use sum aggregation, which is both efficient and expressive in capturing neighbor information. The GIN layer updates node embeddings \( X' \) by combining the self-loop features \( X \) and aggregated neighbor features using the adjacency matrix \( A \):
\[
X' = \epsilon \cdot X + \text{Aggregate}(X_{\text{neighbors}}),
\]
where: \( \epsilon \) is a learnable parameter that scales the self-loop contribution and \( \text{Aggregate}(X_{\text{neighbors}}) = A X \), using the sum aggregation strategy (as before, the identity term is removed due to the residual connection that already captures it).

To enhance the representation power, the aggregated features are processed through an MLP:
\[
\text{MLP}(z) = \text{Linear}_2(\text{Activation}(\text{Norm}(\text{Linear}_1(z)))),
\]
where \( \text{Linear}_1 \) and \( \text{Linear}_2 \) are fully connected layers, with \( \text{Linear}_1 \) projecting the input to a hidden dimension controlled by a multiplier \( \gamma \), and \( \text{Linear}_2 \) mapping it back to the output dimension and \( \text{Norm} \) applies normalization (e.g., LayerNorm), and a ReLU activation function introduces non-linearity.

We enforce causality by applying a mask to edges such that only edges \( (i, j) \) where \( i \leq j \) within the same graph are retained:
\[
\text{Causal Mask: } \text{edge\_index}_{\text{filtered}} = \{(i, j) \mid i \leq j, \text{ and } \text{batch}[i] = \text{batch}[j]\}.
\]
This ensures that nodes only receive information from preceding nodes, maintaining a directed acyclic graph structure.

The final output of the GIN layer is computed by combining the transformed node features with a residual connection:
\[
X_{\text{residual}} = \text{Linear}(X),
\]
\[
X_{\text{final}} = \text{MLP}(X') + X_{\text{residual}}.
\]
This residual connection ensures smooth gradient flow and stabilizes training, particularly when the input and output dimensions differ.

The use of MLPs with normalization and activation enhances the representation learning capabilities, while residual connections ensure robust integration of the GIN layer into the overall architecture. By simplifying the aggregation strategy and emphasizing causality, the approach efficiently captures relational dependencies while adhering to computational constraints.

When used as a fine-tuning strategy, only parameters in this newly added GIN are trainable and the rest of the model is frozen per its pre-trained weights. 

\begin{figure}
	\centering
	\includegraphics[width=0.75\linewidth]{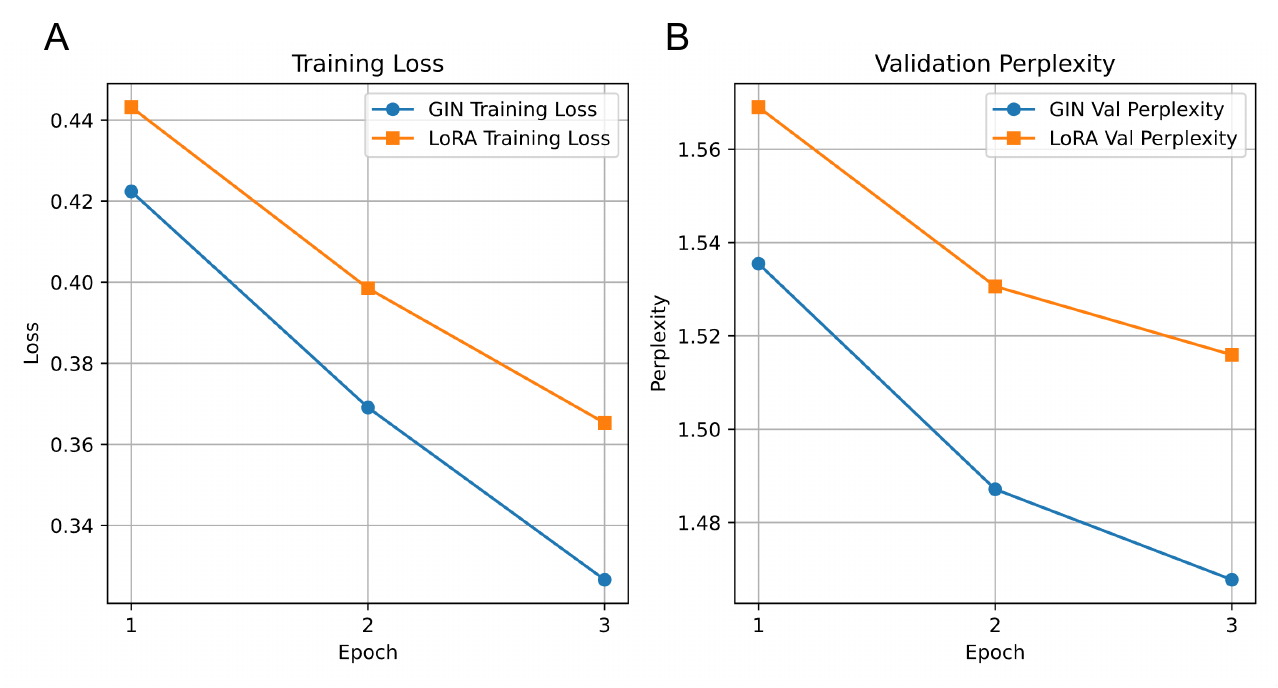}
        \caption{Performance of LoRA fine-tuning (panel A) and sparse GIN fine-tuning. In sparse GIN fine-tuning, we interpret the attention matrix computed by the pre-trained model as an adjacency matrix. Here, we sum attention matrices across all heads and clamp at 1.0, and then use it as an input to a GIN model. Only \hlyellow{adjacency} matrix values above a threshold of 0.2 are considered, introducing a sparseness and computational efficiency. Both LoRA and sparse GIN feature the same number of trainable parameters. Panel A: Training loss over epochs for LoRA and sparse GIN. Sparse GIN demonstrates faster convergence and lower final training loss compared to LoRA, indicating improved optimization efficiency. Panel B: Validation perplexity over epochs for LoRA and sparse GIN. Sparse GIN achieves lower perplexity across all epochs, suggesting better generalization to unseen data.    }
	\label{fig:fig_202}
\end{figure}

Figure~\ref{fig:fig_202} shows the performance of LoRA fine-tuning (Figure~\ref{fig:fig_202}A) and sparse GIN fine-tuning. In sparse GIN fine-tuning, we interpret the attention matrix computed by the pre-trained model as an adjacency matrix. Here, we sum attention matrices across all heads and clamp at 1.0, sparsify the adjacency matrix, and then use it as an input to a GIN model (details, see Materials and Methods). Both LoRA and sparse GIN feature the same number of trainable parameters. Next, Figure~\ref{fig:fig_202}A shows training loss over epochs for LoRA and sparse GIN. S

We find that sparse GIN demonstrates improved convergence and lower final training loss compared to LoRA, indicating improved optimization efficiency and training dynamics. Figure~\ref{fig:fig_202}B visualizes validation perplexity over epochs for LoRA and sparse GIN. We also observe that sparse GIN achieves lower perplexity across all epochs, suggesting better generalization to unseen data. These results are in general agreement with the data obtained for GIN-Attention per the previous section, suggesting that the graph-based approach yields better generalization and less overfitting. Especially the finding that sparse GIN fine-tuning is easy to implement with few parameters, but with significantly improved trainining dynamics, offers interesting future opportunities. 

Another interesting aspect is that we can feed an external adjacency matrix to be added to the values identified by the attention mechanisms; if this is included during training it can provide a straightforward way to incorporate structured graph information into a pre-trained model that was originally not necessarily identified for this objective. 

\begin{figure}
	\centering
	\includegraphics[width=0.75\linewidth]{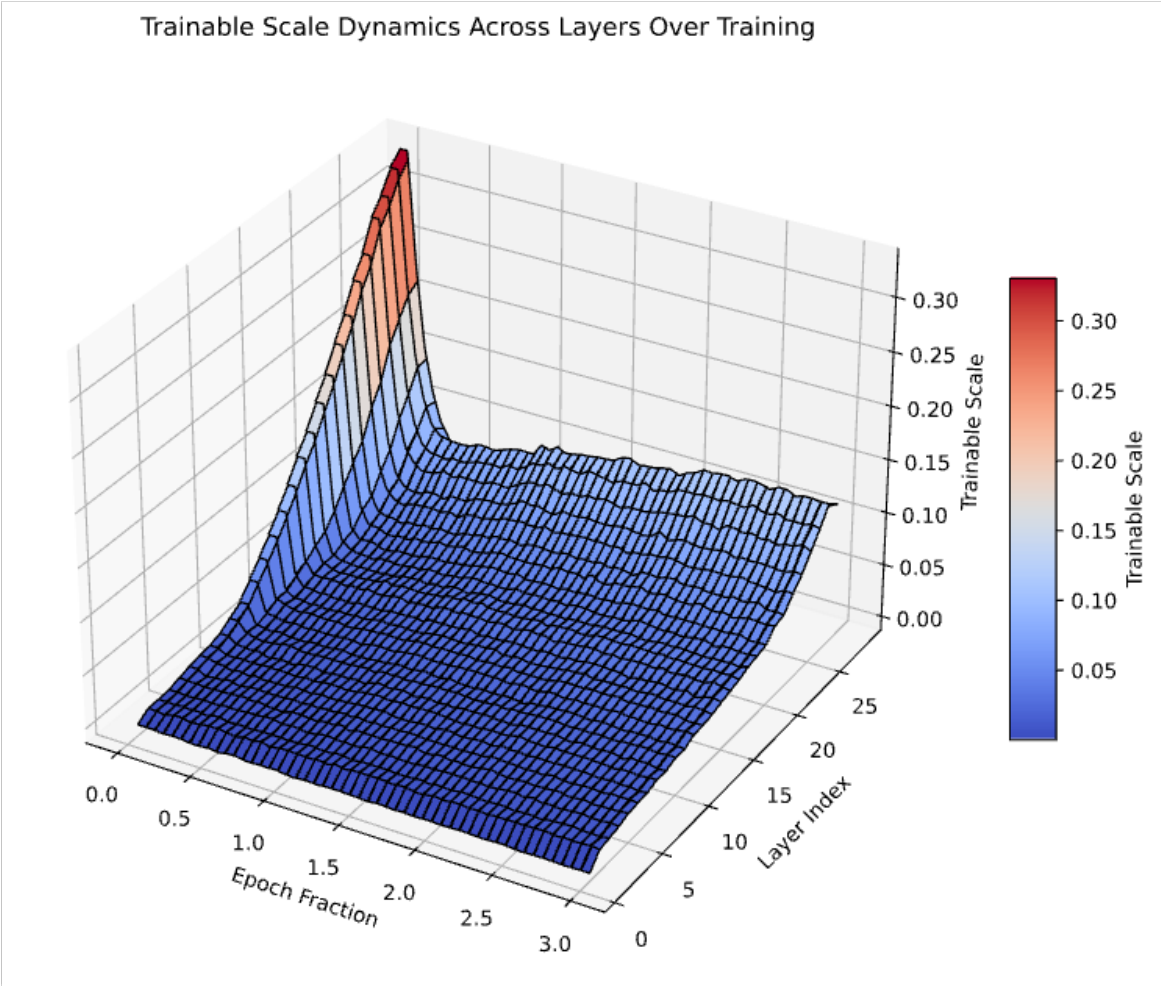}
        \caption{Trainable scale parameter  $\lambda$  over all $k$ layers in the model, plotted over all epochs. The trainable scale parameter delineates the relative importance of the sparse GIN as it is added to the original signal. The plot illustrates how the scale parameter evolves over both the layer index and the epoch fraction. Early in training, higher layers exhibit stronger scaling values, indicating a higher reliance on sparse GIN adjustments. As training progresses, the scaling values stabilize, suggesting convergence in the relative importance of the sparse GIN contributions across layers. The color gradient reflects the magnitude of the scale parameter, with warmer colors (red) indicating higher values and cooler colors (blue) indicating lower values. This visualization provides insights into the adaptive behavior of the trainable scale parameter over the course of training.
  }
	\label{fig:fig_200}
\end{figure}

A key parameter in the sparse GIN model is the use of the trainable scale parameter that delineates how the results of GIN updates are merged with the original signal. Figure~\ref{fig:fig_200} depicts the trainable scale parameter $\lambda$ over all layers in the model, plotted over all training epochs. Early in training, higher layers exhibit stronger scaling values, indicating a higher reliance on sparse GIN adjustments. As training progresses, the scaling values stabilize, demonstrating convergence in the relative importance of the sparse GIN contributions across layers. The color gradient reflects the magnitude of the scale parameter, with warmer colors (red) indicating higher values and cooler colors (blue) indicating lower values. This visualization provides insights into the adaptive behavior of $\lambda$ in each layer over the course of training.

\begin{figure}
	\centering
	\includegraphics[width=1.\linewidth]{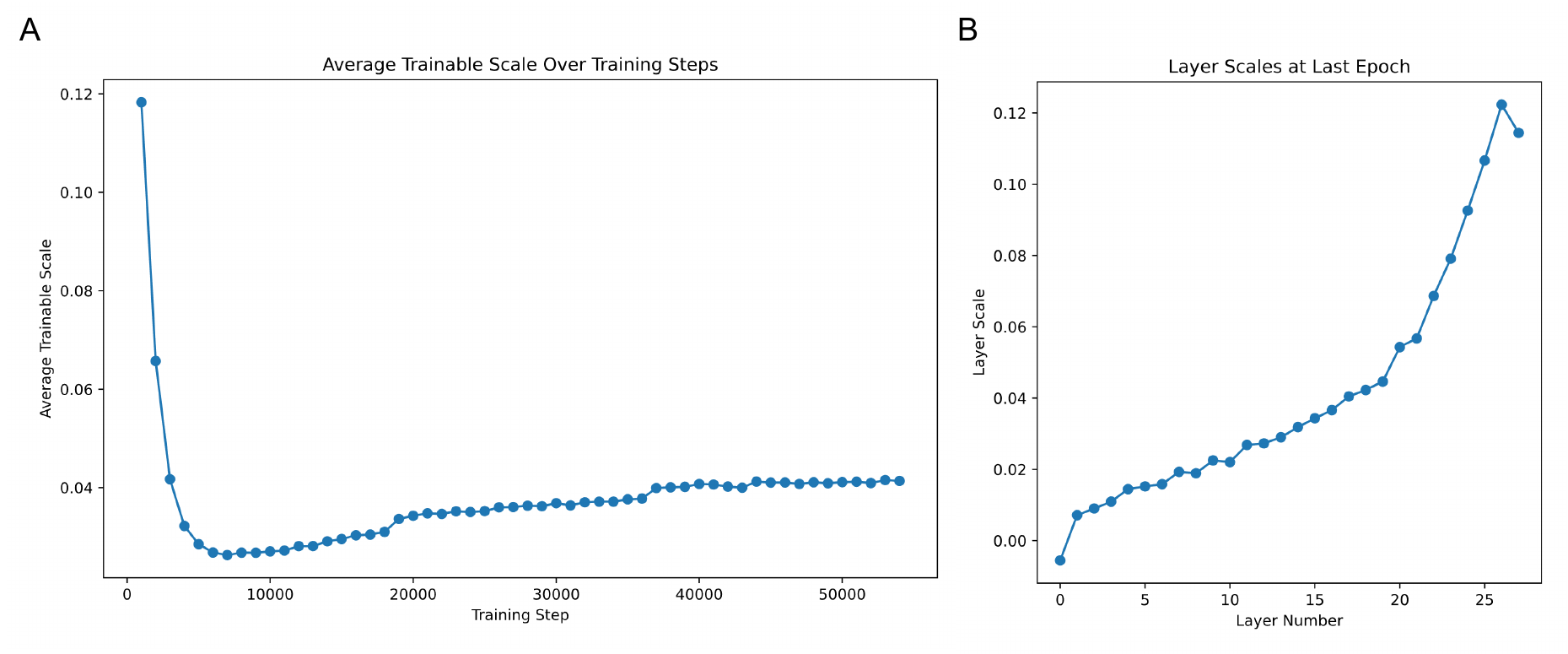}
        \caption{Global dynamics of the trainable scale parameter $\lambda$ during training and across model layers $k$. 
        Panel A visualizes the average trainable scale parameter over training steps. The plot illustrates a rapid decline in the average scale parameter during the initial stages of training, indicating early adaptation of the sparse GIN contributions. After the initial drop, the scale stabilizes and gradually increases slightly, suggesting the model fine-tunes the integration of sparse GIN as training progresses.  Panel B displays the trainable scale parameter for each layer at the last epoch. The scale parameter exhibits an increasing trend from lower to higher layers, reflecting the progressively stronger reliance on sparse GIN in deeper layers of the model. This layer-wise scaling suggests that deeper layers benefit more from the structural adjustments provided by sparse GIN.}
	\label{fig:fig_201}
\end{figure}

For a more global perspective on how $\lambda$ behaves during training, Figure~\ref{fig:fig_201} shows two analyses. Figure~\ref{fig:fig_201}A depicts the average trainable scale parameter over training steps, revealing a rapid decline in the average scale parameter during the initial stages of training, indicating early adaptation of the sparse GIN contributions. After the initial drop, the scale stabilizes and gradually increases slightly, suggesting the model fine-tunes the integration of sparse GIN as training progresses.  Figure~\ref{fig:fig_201}B displays the trainable scale parameter for each layer after the last training epoch. The scale parameter exhibits an increasing trend from lower to higher layers, reflecting the progressively stronger reliance on sparse GIN in deeper layers of the model. We note that this layer-wise scaling suggests that deeper layers benefit more from the structural adjustments provided by sparse GIN (this observation informed our choice for the initial depth scaling of the parameter as defined above).

\begin{figure}
	\centering
	\includegraphics[width=0.75\linewidth]{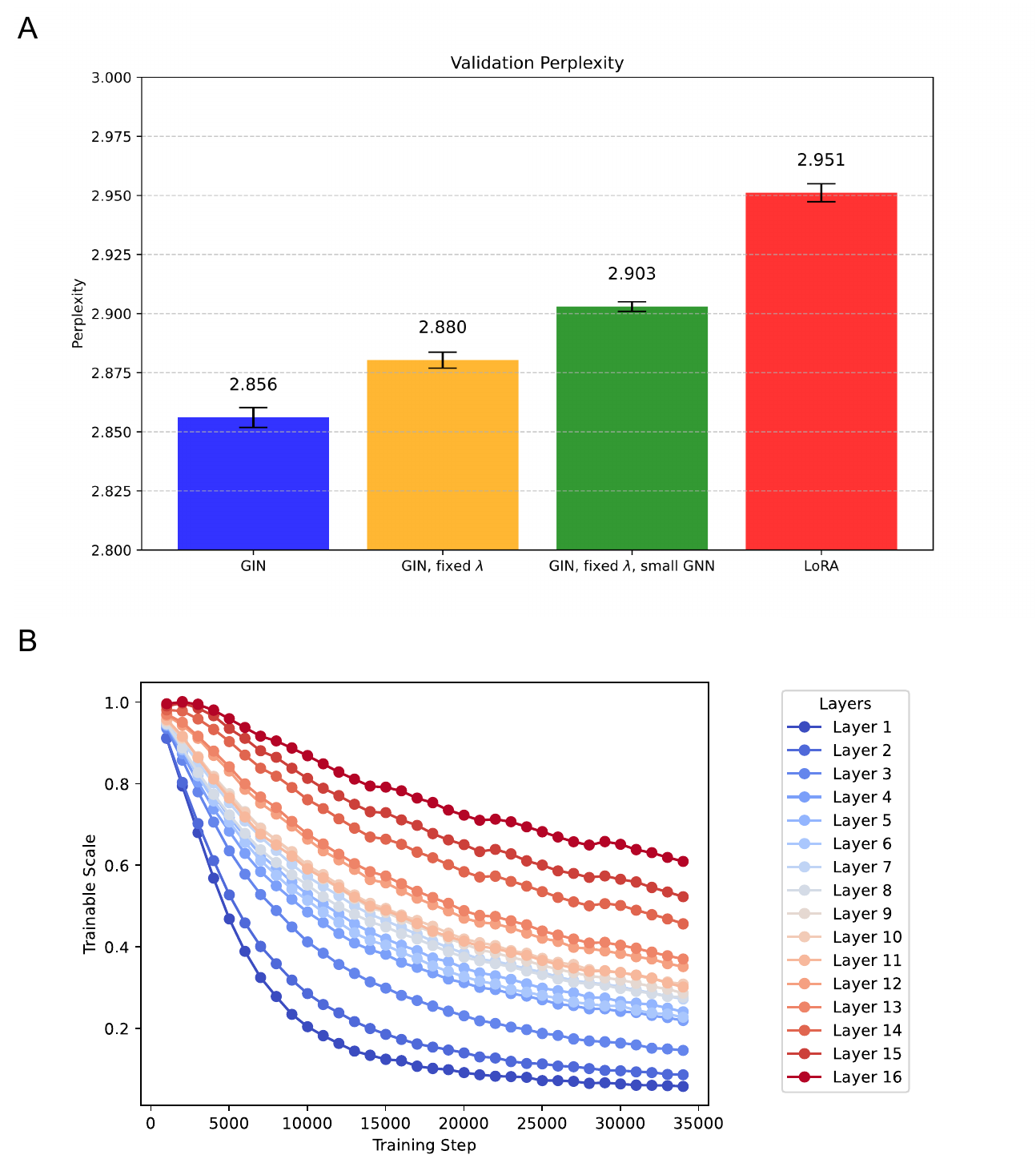}
        \caption{Validation perplexity comparison between different model configurations. Panel A: The bar plot illustrates the validation perplexity values for GIN, GIN with fixed $\lambda$, GIN with fixed $\lambda$ and a smaller GNN, and LoRA. Measured values and error bars represent the standard deviation of the measured perplexity in the last training epoch. GIN achieves the lowest perplexity, while LoRA exhibits the highest perplexity. Panel B shows the trainable scale parameter $\lambda$. Similar to the earlier results, the parameter is found to ultimately be smallest in earlier layers, and largest in deep layers. }
	\label{fig:fig_203}
\end{figure}

In another experiment we trained a Sparse-GIN model on a symbolic reasoning dataset to test how general the performance improvements are, along with a few additional variations (see, Figure~\ref{fig:fig_203}). 
The results of the validation perplexity analysis for various model configurations are presented in Figure~\ref{fig:fig_203}. Figure~\ref{fig:fig_203}A highlights the performance differences between GIN, GIN with fixed $\lambda$, GIN with fixed $\lambda$ and a smaller GNN, and LoRA. As evident from the results, the Sparse-GIN model with learnable $\lambda$ achieves the lowest validation perplexity (2.856), outperforming all other configurations. This indicates that GIN approach effectively captures the necessary relationships in the data. The Sparse-GIN with fixed $\lambda$ configuration slightly increases the perplexity, suggesting that fixing $\lambda$ limits the flexibility of the model to adapt to data. Using a smaller GNN in the GIN with fixed $\lambda$ configuration (GIN, fixed $\lambda$, small GNN) results in slightly worse performance. This increase implies that the smaller GNN size might reduce the model's capacity to fully capture intricate dependencies, thus leading to a minor degradation in performance. The LoRA model exhibits the highest perplexity across all cases, which may indicate that while LoRA introduces additional parameter-efficient mechanisms, it does not optimize as effectively for this specific task. The error bars in the figure represent the standard deviations across multiple runs, confirming the statistical significance of the observed differences. These findings underscore the superior performance of the GIN model in minimizing validation perplexity and highlight the trade-offs introduced by modifying $\lambda$ or reducing the GNN size. Such insights are valuable for designing future architectures to balance efficiency and performance. Figure~\ref{fig:fig_203}B shows the trainable scale parameter $\lambda$. Similar to the earlier results, the parameter is found to be smallest in earlier layers, and largest in deep layers. 

\section{Conclusion}

We explored theoretical and practical adaptations of the Transformer architecture through the lens of graph-based models and category theory. By interpreting and formulating Transformers as Graph Isomorphism Networks (GIN), we established a novel perspective on their attention mechanisms, demonstrating structural equivalence between the aggregation processes in graph neural networks and the focus mechanisms in attention layers. This equivalence provides a foundation for extending Transformers with explicit graph reasoning capabilities that \hlyellow{manifest themselves} in better generalization performance.

We further explored the use of Principal Neighborhood Aggregation (PNA) method, which augments attention by integrating multiple aggregation techniques per attention head, testing more diverse representations. While performance did not exceed that of the GIN model, this enhancement, coupled with the flexibility of graph neural networks, could offer an alternative framework for handling structured and relational data in domains such as bioinformatics, materials science, and other areas.

From a theoretical standpoint, we incorporated insights from category theory to frame Transformers as functors that map input to output representations while preserving structural relationships. This abstraction highlights their inherent capacity for compositional reasoning and contextual adaptation, bridging concepts from information theory, statistical physics, and graph theory. Additionally, the categorical perspective inspires future directions, such as exploring natural transformations for designing more robust and adaptive attention mechanisms that incorporate powerful graph modeling techniques like GINs or PNAs.

Our proposed sparse GIN fine-tuning approach can be used as an alternative or complement to LoRA~\cite{Hu2021LoRA:Models}, using sparse adjacency matrices derived from attention mechanisms as inputs to graph neural networks. This method significantly enhances the model's ability to learn from mathematical training data, demonstrating improved performance and flexibility. We find that Sparse-GIN with learnable $\lambda$ achieves the lowest validation perplexity, demonstrating superior adaptability and performance, while fixed $\lambda$, smaller GNN size, and LoRA configurations show trade-offs in flexibility and capacity (Figure~\ref{fig:fig_203}). These deeper insights provide additional evidence for the findings, especially since they hold for a different dataset. 

The findings and methods presented in this work not only deepen the understanding of attention mechanisms but also pave the way for innovative applications of Transformers integrated with graph-based reasoning. By synthesizing theoretical rigor with practical innovation, this work contributes to the ongoing evolution of neural network architectures, advancing their potential to address increasingly complex challenges in science and technology.

\subsection{Transformers are secretly GIN-like graph reasoning models}

The discussion in this section explains how a standard transformer architecture can be viewed as a GIN-type graph neural network. We posit that vanilla Transformer architectures (as shown in Figure~\ref{fig:fig_775}A) are, in fact, GIN-like graph models. It thereby invokes our modeling strategy as a ``GIN-within-GIN'' mechanism, which nests per-head GINs inside a global GIN, yielding a hierarchical model for graph-structured data.

A transformer layer consists of multi-head self-attention. For each head $h = 1, 2, \dots, H$, we compute:
\[
\mathbf{A}_h
= \mathrm{softmax}\!\biggl(\frac{\mathbf{Q}_h \,\mathbf{K}_h^\top}{\sqrt{d_k}}\biggr),
\quad
\mathbf{O}_h
= \mathbf{A}_h \,\mathbf{V}_h,
\]
where
\begin{itemize}
    \item $\mathbf{Q}_h, \mathbf{K}_h, \mathbf{V}_h \in \mathbb{R}^{T \times d_k}$ are the query, key, and value matrices for head $h$;
    \item $T$ is the (sequence or node) dimension;
    \item $d_k$ is the per-head feature dimension.
\end{itemize}

These quantities are computed as follows: 
\[
\mathbf{Q}_h = \mathbf{H} \mathbf{W}_h^Q,
\quad
\mathbf{K}_h = \mathbf{H} \mathbf{W}_h^K,
\quad
\mathbf{V}_h = \mathbf{H} \mathbf{W}_h^V,
\]
where:

\begin{itemize}
    \item \( \mathbf{H} \in \mathbb{R}^{T \times d_k} \): The input hidden states, where \( T \) is the sequence length (or number of nodes) and \( d \) is the hidden dimension.
    \item \( \mathbf{W}_h^Q, \mathbf{W}_h^K, \mathbf{W}_h^V \in \mathbb{R}^{d \times d_k} \): Learnable weight matrices specific to head \( h \), projecting the input hidden states into the query, key, and value spaces, respectively.
    \item \( \mathbf{Q}_h, \mathbf{K}_h, \mathbf{V}_h \in \mathbb{R}^{T \times d_k} \): The query, key, and value matrices for head \( h \), where \( d_k = d / H \) is the feature dimension for each head.
\end{itemize}

These projections transform the input into three distinct spaces:
\begin{itemize}
    \item The \textbf{query} matrix \( \mathbf{Q}_h \) captures the representation of the tokens in terms of the features they query.
    \item The \textbf{key} matrix \( \mathbf{K}_h \) represents the features available for attention.
    \item The \textbf{value} matrix \( \mathbf{V}_h \) provides the features that will be aggregated based on the attention weights.
\end{itemize}

Each head's output $\mathbf{O}_h \in \mathbb{R}^{T \times d_k}$ is concatenated along the feature dimension:
\[
\mathbf{H}_{\mathrm{attn}}
 =\bigoplus_{h=1}^H \mathbf{O}_h,
\]
resulting in a $\mathbb{R}^{T \times (H\,d_k)}$ matrix. A shared projection reduces this to the original hidden dimension $d$:
\[
\mathbf{H}_{\mathrm{attn}}^{(k)}
= \mathbf{H}_{\mathrm{attn}} \,\mathbf{W}^O,
\]
where $\mathbf{W}^O \in \mathbb{R}^{(H\,d_k) \times d}$ is learnable.

Each attention head $\mathbf{A}_h \in \mathbb{R}^{T \times T}$ serves as an adjacency matrix, specifying how each ``node'' (token) attends to others. This adjacency-guided aggregation is central to the connection between transformers and GINs.

In a Graph Isomorphism Network (GIN), the layer update for a node $v$ at the $k$-th layer is:
\[
\mathbf{h}_v^{(k)}
= \mathrm{MLP}^{(k)}\!\Bigl(
(1 + \epsilon)\,\mathbf{h}_v^{(k)}
\;+\;\sum_{u \,\in\, \mathcal{N}(v)} \mathbf{h}_u^{(k)}
\Bigr),
\]
where $\epsilon$ is a scalar controlling the self-contribution, and $\mathrm{MLP}$ is a multi-layer perceptron for feature transformation.

Similarly, in a transformer, the self-attention mechanism computes:
\[
\mathbf{H}_{\mathrm{attn}}^{(k)}
= \mathbf{H}^{(k)}
+ \Bigl(\sum_{h=1}^H \mathbf{A}_h \, \bigl(\mathbf{H}^{(k)} \mathbf{V}_h^{(k)}\bigr)\Bigr) \mathbf{W}_h^{(k),O}
\]
where $\mathbf{A}_h$ is the adjacency matrix from attention, $\mathbf{W}_h^{(k),O}$ is the projection matrix, and the residual connection ensures self-contribution.

The resemblance becomes complete with the application of the feed-forward network, which contains an MLP:
\begin{equation}
\mathbf{H}^{(k)}
=\mathrm{MLP}\bigl(\mathbf{H}_{\mathrm{attn}}^{(k)}\bigr) =  \mathrm{MLP}\bigl( \mathbf{H}^{(k)}
+ \Bigl(\sum_{h=1}^H \mathbf{A}_h \,\mathbf{H}^{(k)} \Bigr)\,\mathbf{W}_h^{(k),O} \bigr).
\end{equation}
Here:
\begin{itemize}
    \item The attention mechanism aggregates information from neighbors, here through a multi-headed process (like $\sum_{u \in \mathcal{N}(v)}$ in GINs).
    \item The FFN applies a non-linear transformation, analogous to $\mathrm{MLP}$ in GINs.
    \item The residual connection corresponds to the $(1 + \epsilon)$ term in GINs, with $\epsilon=0$ in transformers.
\end{itemize}

Transformers hence resemble a GIN-like process, where attention handles neighbor aggregation, and the feed-forward MLP handles nonlinear feature transformation. There are, of course, notable distinctions. For instance, Transformers concatenate the outputs of multiple heads and then apply a single projection and feed-forward layer, rather than summing neighbor features and applying an MLP per node. In GINs, the MLP directly follows the aggregated representation of each node and its neighbors (scaled by \( (1 + \epsilon) \)). Transformers, however, first mix token embeddings across multiple attention heads (via concatenation) before applying a global linear projection \( \mathbf{W}_o \). Only after this step does the transformer’s feed-forward network come into play. These differences in adjacency construction, concatenation versus summation, and MLP placement mean that transformers and GINs, while sharing certain high-level motifs, are architecturally distinct and exhibit different inductive biases. Our GIN-within-GIN model addresses these differences and formally adds hierarchical GINs for more expressive graph processing with proper injective properties.  

Our GIN-Attention architecture adds a hierarchical structure by nesting per-head GINs inside a global GIN-like algorithm. Let $\mathbf{H}^{(k)} \in \mathbb{R}^{T \times d}$ be the input node-feature matrix for the $k$-th layer. We define:
\begin{equation}
\mathbf{H}^{(k)}
= \mathrm{MLP}_{\mathrm{global}}^{(k)}\!\Bigl(\mathbf{H}^{(k)}
+ \Bigl( \bigoplus_{h=1}^H
\mathrm{MLP}_h^{(k)}\!\Bigl(
(1 + \epsilon)\mathbf{H}^{(k)}
+ \mathbf{A}_h^{(k)} \,\mathbf{H}^{(k)}
\Bigr)
\Bigr)
\,\mathbf{W}_{O}^{(k)}\,\Bigr)
\end{equation}
where:
\begin{itemize}
    \item $\mathrm{MLP}_h^{(k)}(\cdot)$ is an independent GIN graph neural network for head $h$.
    \item $\mathbf{A}_h^{(k)}$ is the adjacency (attention) matrix from head $h$.
    \item $\mathbf{W}_{O}^{(k)}$ is an optional learnable projection matrix applied after concatenation across heads and global GIN aggregation.
    \item $\mathrm{MLP}_{\mathrm{global}}^{(k)}(\cdot)$ integrates the outputs of all per-head aggregated features to produce a global representation.
\end{itemize}
Each head specializes in capturing localized patterns using its own GIN, and the global GIN integrates these representations for the next layer.

By nesting GINs within each attention head, we obtain more powerful hierarchical model for graph relational modeling. As our results showed, this increases performance at the same parameter count and specifically with increased generalization capabilities and reduced overfitting. 

These and other insights complement other works where researchers have elucidated important principles and strategies implicitly or tacitly captured in the original transformer architecture, providing increasing explanation for the broad usability and impact the model architecture has been able to achieve~\cite{teo2024unveilinghiddenstructureselfattention,razzhigaev2024transformersecretlylinear,oren2024transformersmultistaternns,pfau2024letsthinkdotdot,sanford2024understandingtransformerreasoningcapabilities,wei2023chainofthoughtpromptingelicitsreasoning,li2024chainthoughtempowerstransformers,zhang2024differentstructuralsimilaritiesdifferences,ruoss2024amortizedplanninglargescaletransformers}. We hope that our initial results and strategy presented here can be explore in other models, with other data sets and tasks, and tested in other settings.

\subsection{Theoretical foundations of sparsification in GNN-Based transformers}
 
Sparsification in GNN-based Transformers introduces a principled approach to improving scalability and interpretability while enhancing the relational reasoning capabilities of attention mechanisms. At its core, sparsification reflects a foundational shift from dense, exhaustive relational modeling to selective, task-relevant graph operations. This aligns with real-world data distributions, which are often sparse and structured (e.g., social networks, molecular graphs, dependency trees in language).

Graph-based neural networks and attention mechanisms are inherently complementary: Attention dynamically determines relevance among tokens, while GNNs encode structural relationships through adjacency matrices. Sparsification introduces an explicit relational inductive bias by filtering connections in the graph, focusing on task-relevant edges while discarding noise. This selective process reflects principles from graph sparsity theory, where retaining high-weight edges preserves the graph's expressive power~\cite{xu2019how, corso2020principal, spielman2011graph}. 

The \hlyellow{aggregated}, sparsified matrix $A'$ defined in Section~\ref{sparse-gin-formalism} retains the most significant relationships across attention heads, aligning the computation with the graph sparsity principles observed in real-world relational data. This focus on sparse, high-weight edges not only improves computational efficiency but also introduces regularization, mitigating overfitting in high-dimensional spaces.

The integration of sparsification into multi-head attention transforms each Transformer layer into a hierarchical graph reasoning module. By aggregating information across all attention heads into a single effective adjacency matrix \( A'^{(k)} \) at layer $k$, the framework models complex relationships while preserving scalability. Node embeddings \( H^{(k)} \) are updated as:
\begin{equation}
\label{hierarc_GIN_eq}
H^{(k)} = \text{MLP}^{(k)}_{\text{global}}\left(H^{(k-1)} + \lambda^{(k)}\text{MLP}^{(k)}_{\text{Sparse-GIN}}\left((1 + \epsilon)H^{(k-1)} + A'^{(k)} H^{(k-1)}\right)\right),
\end{equation}
where \( \epsilon \) controls the self-loop contribution and $\lambda^{(k)}$ is the trainable scale parameter. This hierarchical reasoning enables each layer to capture global and local dependencies while adapting to sparse relational structures.

\paragraph{Compositionality in Category Theory} The Sparse GIN framework can be interpreted as a compositional system for relational reasoning, rooted in the principles of category theory~\cite{Eilenberg1942GroupHomology,Eilenberg1945GeneralEquivalences,Giesa2012CategoryDesign,Spivak2011CategoryNetworks}. In this view, the Transformer is modeled as a functor \( F: \mathcal{C} \to \mathcal{D} \), mapping objects and morphisms from an input category \( \mathcal{C} \) (e.g., sequences or graphs) to an output category \( \mathcal{D} \) (e.g., embeddings or predictions), while preserving their structural relationships. Sparsification introduces a natural transformation \( \eta: F \to F' \), where \( F' \) represents the sparsified counterpart of the Transformer. Formally, for each input \( x \in \mathcal{C} \), the natural transformation \( \eta \) satisfies:
\[
\eta_x: F(x) \to F'(x),
\]
where \( \eta_x \) modifies the adjacency structure dynamically by transitioning from dense attention to sparsified graph representations. This transition ensures that essential relational information is retained while introducing computational and structural efficiency. Furthermore, the sparsified adjacency matrix \( A_{\text{eff}} \) can be viewed as an enriched morphism in \( \mathcal{C} \), capturing weighted relationships between nodes. By applying \( A'^{(k)} = f\left(\sum_{h=1}^{H} A_h^{(k)}\right) \), the framework dynamically aligns relational reasoning with the task requirements, ensuring compositional consistency across layers. This perspective establishes Sparse-GIN as a modular and mathematically principled extension of Transformers for graph-based reasoning.

\paragraph{Information Bottleneck Theory}  Sparse GIN can be framed through the lens of Information Bottleneck Theory,~\cite{tishby2000information,saxe2018information} which posits that a model should retain only the most relevant information from the input while discarding redundant or noisy details. In this context, sparsification introduces an information bottleneck by filtering low-weight edges in the adjacency matrix \( A'^{(k)} \), ensuring that only task-critical relationships are preserved. Formally, the sparsified matrix \( A'^{(k)} \) can be seen as an intermediate representation \( Z \) that minimizes the mutual information \( I(X; Z) \) between the input \( X \) and the sparsified representation, subject to a constraint on preserving the predictive information \( I(Z; Y) \) necessary for the target output \( Y \):
\[
\min_{A'} I(X; Z) \quad \text{subject to} \quad I(Z; Y) \geq \text{threshold}.
\]
This trade-off may enable Sparse-GIN to reduce overfitting by discarding irrelevant features while maintaining expressive power. By modeling sparsification as an information bottleneck, the framework achieves robust generalization across diverse tasks while aligning with real-world graph sparsity observed in domains like molecular interactions,  mechanistic principles in physics, or social networks.

\paragraph{Hierarchical Representation Learning} Sparse GIN may naturally support hierarchical representation learning by aggregating information across local and global relational structures within graphs~\cite{hamilton2018representationlearninggraphsmethods}. Each Transformer layer \( k \) models node-level interactions using the sparsified adjacency matrix \( A'^{(k)} \), while progressively building higher-order abstractions through inter-layer aggregation. As defined in equation~(\ref{hierarc_GIN_eq}), the node embeddings \( H^{(k)} \) in each layer are updated to selectively retain the most significant relationships, ensuring task-relevant features are propagated. To capture global hierarchical dependencies, the final representation aggregates features across layers using the FF MLP. This hierarchical structure enables Sparse-GIN to model complex dependencies at multiple levels of abstraction, bridging local interactions and global task objectives. This multi-level reasoning aligns with cognitive processes and supports applications requiring deep relational insights, such as molecular property prediction and symbolic reasoning.

\paragraph{Summary of implications}
The combination of sparsification and hierarchical graph reasoning provides a framework bridges structured and sequential data modeling, and could be useful as a fact of scalable and interpretable foundation for AI systems. These principles lay the groundwork for future advancements in sparsity-aware attention mechanisms and their applications to diverse domains such as bioinformatics, symbolic reasoning, and multi-modal learning. We find that Sparse-GIN demonstrates robust generalization across diverse fine-tuning experiments, including mathematical reasoning tasks, and symbolic logic datasets, achieving strong performance across all domains.

The sparsification of adjacency matrices in GIN-based Transformers carries foundational implications for structured and sequential data modeling:
\begin{itemize}
    \item \textbf{Scalability}: By reducing computational complexity, sparsification extends the applicability of Transformers to large graphs and long sequences.
    \item \textbf{Robustness and Regularization}: Sparsity acts as an implicit regularizer, improving generalization and mitigating overfitting in high-dimensional spaces.
    \item \textbf{Alignment with Real-World Data}: Many real-world graphs, such as molecular structures or dependency graphs, exhibit sparsity. By modeling these distributions explicitly, the framework naturally aligns with the underlying structure of the data.
    \item \textbf{Expressive Power}: Sparsified graph reasoning retains essential structural information while discarding redundant or noisy edges, maintaining expressivity with reduced complexity.
    \item \textbf{Auxiliary Control with External Adjacency Matrices}: 
    The framework supports the integration of external adjacency matrices during training and inference, enabling auxiliary control over the relational structure. For example, domain-specific knowledge encoded as a predefined adjacency matrix can guide the model to focus on specific dependencies, such as known molecular interactions or pre-defined dependency structures. Future experiments could explore scenarios where external adjacency matrices are introduced only during inference to adapt the model dynamically to new contexts.
    \item \textbf{New Foundation Model Development and Training Strategy}: 
    This approach, essentially implemented in our fine-tuning experiments, suggests a novel training paradigm: first, pretrain and even fine-tune a standard Transformer on the target task or a related dataset, and then augment it with Sparse-GIN as an additional strengthening mechanism. By retaining the pretrained Transformer’s core capabilities and introducing sparsified graph reasoning in a second stage, this method combines the benefits of dense pretraining with sparse fine-tuning, enabling task-specific enhancements without retraining the entire model. As an alternative, one could also train a GIN-Attention model and then add a Sparse-GIN layer on top of it.  
\end{itemize}

\subsection{Broader perspective and outlook}
 
A key objective of AI for science is the advancement of discovery via generalization, moving from fitting to training data towards more expressive, generalizable AI systems that can reason over never-before-seen data and problems and understand principles to obtain solutions rather than memorizing answer. When considered in the context of biological problems, for instance, the concept of GIN-Attention, which integrates Graph Isomorphism Networks (GINs) into Transformer attention mechanisms, aligns seamlessly with the principles of materiomics~\cite{Cranford2012Biomateriomics,Shen2022ComputationalMateriomics,BuehlerMSME_BuildBlock_2023,Guo2020ANetworks,Guo2022RapidNetworks,reiser2022gnn_materials}, the holistic study of material properties and behaviors through their intrinsic structures and interactions. In materiomics, understanding the hierarchical and relational nature of molecular or microstructural patterns is crucial, as these relationships dictate material performance across scales. These are powerful principles in biological materials and living organisms~\cite{Buehler2023GenerativeDesign,Arevalo2023LearningMaterials}, and are key to complex problem-solving in biological agentic systems~\cite{doi:10.1073/pnas.2414274121}. 

\begin{figure}
\sffamily 
\small
\sansmath 

\begin{tikzpicture}[node distance=1.5cm]

\node (start) [startstop] {Material Microstructure \\ (e.g., Proteins, Polymers)};
\node (graph1) [graph, below of=start] {Graph Representation \\ (Initial Layer, Layer 1)};
\node (attention1) [process, below of=graph1, xshift=-2cm] {Attention  \\ (Layer 1, Head 1)};
\node (attention2) [process, below of=graph1, xshift=2cm] {Attention  \\ (Layer 1, Head 2)};
\node (merge1) [graph, below of=attention1, yshift=-0cm, xshift=2cm] {Updated Graph \\ Representation};

\node (graph2) [graph, below of=merge1, yshift=-.5cm] {Graph Representation \\ (Layer 2)};
\node (attention3) [process, below of=graph2, xshift=-2cm] {Attention  \\ (Layer 2, Head 1)};
\node (attention4) [process, below of=graph2, xshift=2cm] {Attention  \\ (Layer 2, Head 2)};
\node (merge2) [graph, below of=attention3, yshift=-0cm, xshift=2cm] {Final Learned \\ Representation};

\node (insights) [result_rounded, below of=merge2] {Materiomic Task Solution \\ (e.g. Material Properties, Inverse Design, ...)};

\draw [arrow] (start) -- (graph1);
\draw [arrow] (graph1) -- (attention1);
\draw [arrow] (graph1) -- (attention2);
\draw [arrow] (attention1) -- (merge1);
\draw [arrow] (attention2) -- (merge1);
\draw [arrow] (merge1) -- (graph2);
\draw [arrow] (graph2) -- (attention3);
\draw [arrow] (graph2) -- (attention4);
\draw [arrow] (attention3) -- (merge2);
\draw [arrow] (attention4) -- (merge2);
\draw [arrow] (merge2) -- (insights);

\node[align=left, left of=graph1, xshift=-4.5cm, text width=5cm] (input) {Input: Microstructure features \\ (e.g., atoms, residues, chains)};
\node[align=left, left of=attention1, xshift=-4.5cm, text width=5cm] (update) {Each attention head builds \\ a new graph  (edges updated per adjacency matric for each head).};
\node[align=left, left of=merge2, xshift=-4.5cm, text width=5cm] (output) {Final learned representation \\ integrates multi-head updates \\ over all layers.};

\draw [dashed] (input) -- (graph1);
\draw [dashed] (update) -- (attention1);
\draw [dashed] (output) -- (merge2);

\end{tikzpicture}

\caption{Dynamic Graph Representation Learning with GIN-Attention in Transformers
This schematic illustrates the iterative process of GIN-Attention in a Transformer architecture, applied to material microstructures, here conceptually shown for a model with two layers and two heads. Starting with raw microstructural data (e.g., proteins or polymers), an initial graph representation is constructed. At each layer, multiple attention heads dynamically build and refine graph structures by updating adjacency matrices based on learned attention scores. The outputs of all heads are merged to produce updated graph representations, which are iteratively refined across layers. The final learned representation integrates structural and relational insights, enabling the model to predict material properties, uncover structure-property relationships, and design novel materials. This framework highlights the simultaneous graph construction and feature learning facilitated by GIN-Attention.}
\label{fig:materiomics-chart}

\end{figure}
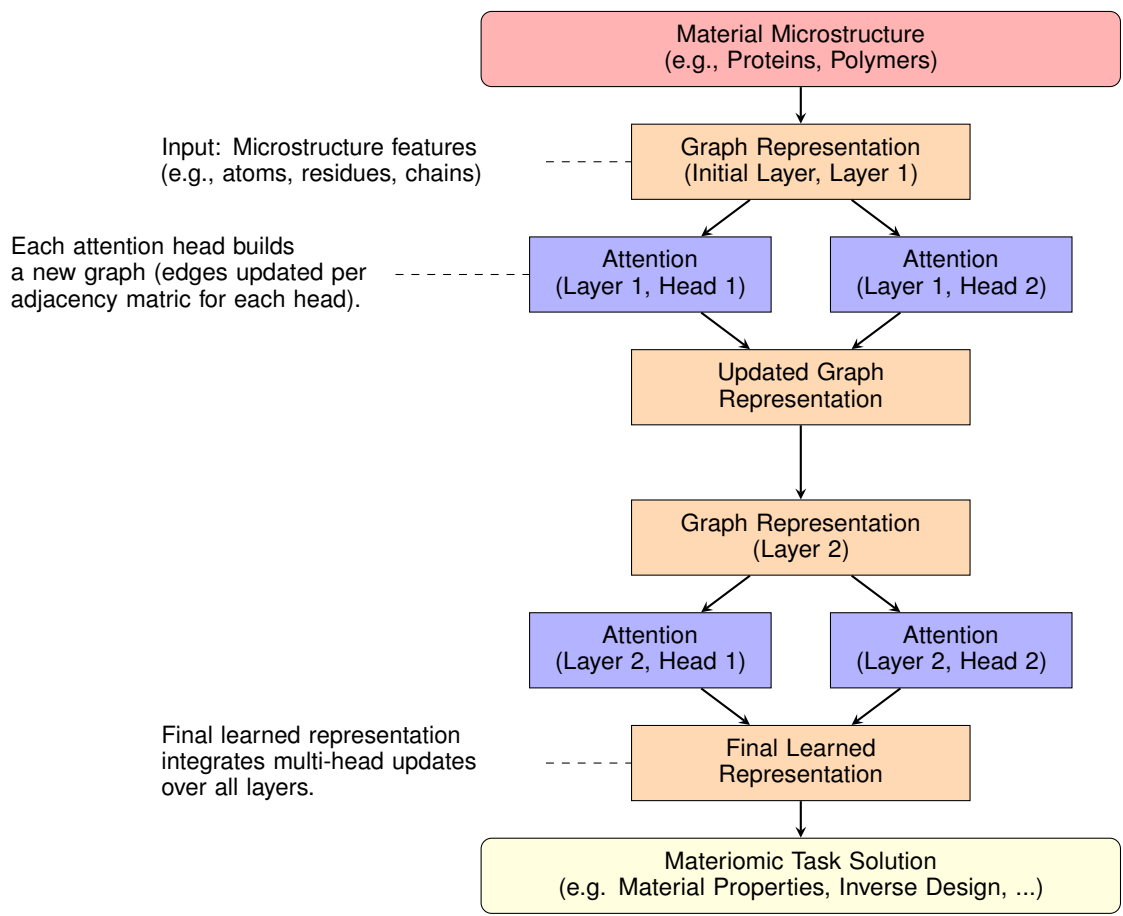

GIN-Attention models these relationships explicitly by interpreting attention weights as graph adjacency matrices, enabling the Transformer to capture intricate dependencies inherent in materials science. For example, in studying protein folding, peptide assembly, the structure and mechanics of polymer chains, or poly-/crystalline lattices, GIN-Attention can enhance the representation of how individual components interact within a material’s architecture, in distinct and varied feature representations and abstractions that emergec naturally at each head and layer (Figure~\ref{fig:materiomics-chart}). By leveraging the graph isomorphism property, it ensures that structurally equivalent elements yield consistent predictions, regardless of orientation or representation, a critical aspect in analyzing materials with symmetrical or repetitive motifs and generalizable principles. This approach opens avenues for more interpretable and precise modeling in materiomics, enabling the design of bioinspired or architected materials by uncovering hidden correlations between microstructure and macroscopic properties. Thus, GIN-Attention bridges advanced machine learning techniques with the foundational challenges of materials science, paving the way for transformative discoveries in the field. Another interesting direction would be to explore how GIN-Attention based models could be integrated into multi-agent systems to further their emergent capabilities and especially scientific discovery~\cite{ghafarollahi2024sciagentsautomatingscientificdiscovery}. 

There are many other fields where such approaches can be useful, in diverse areas ranging from music (where an orchestral piece can be modeled as a complex graph of interactions between notes, instruments, themes, arrangements, and so on, at different scales and resolutions) to protein modeling, to disease, and perhaps even human creativity. We believe that achieving these integrative goals requires bridging graph theory and Transformer architectures, providing injective biases aligned with the implicit graph-like nature of attention mechanisms, and introducing novel methods to enhance their capabilities to relate, reason and develop functors across domains. The integration of Graph-Aware Isomorphic Attention and sparse graph neural network fine-tuning opens new pathways for improving relational reasoning and generalization in Transformers.

\hlyellow{The approach also has substantial implications for materials science, particularly in modeling complex hierarchical systems where structural and functional properties emerge across multiple scales}~\cite{Nepal2022HierarchicallyNanocomposites,Hu2023DeepScience,buehler2024cephalomultimodalvisionlanguagemodels,Buehler2023PredictingCapacityb}. \hlyellow{Materials such as biomaterials, composites, and energy-efficient architectures inherently possess hierarchical and graph-like relational structures, making them ideal candidates for leveraging the enhanced relational reasoning capabilities offered by GIN-Attention. In biomaterials, hierarchical structures spanning molecular interactions up to macroscopic features critically determine mechanical and functional properties}~\cite{Arevalo2023LearningMaterials}. \hlyellow{GIN-Attention’s ability to systematically capture relational dependencies at multiple scales can substantially improve predictive accuracy and generalization for tasks that link microstructural arrangements to macroscale properties like toughness, elasticity, or self-healing capabilities. Composite materials, characterized by intricate interactions among diverse constituent phases, similarly benefit from the Sparse-GIN-Attention mechanism, efficiently identifying and emphasizing sparse yet critical interactions governing their performance. Additionally, designing energy-efficient architectures such as hierarchical metamaterials or lattice structures relies heavily on accurately modeling structural dependencies across scales. The graph-aware attention mechanisms we propose can dynamically pinpoint and leverage relevant structural relationships, thus aiding in discovering optimized architectures for enhanced energy dissipation or load-bearing capacities. Hence, beyond demonstrating general methodological efficacy, applying our approach to materials science underscores practical value, facilitating advancements in computational materials design, property prediction, and accelerated discovery of next-generation materials.}

Looking forward, several opportunities emerge:

\begin{enumerate}
    \item \textbf{Multi-Modal and Cross-Domain Applications}:
    The proposed methods are particularly promising for fields like bioinformatics, materials science, genomics, ecosystems, or systems biology, where both relational and sequential data are prevalent. Extending this approach to multi-modal datasets (e.g., combining images and text) could further enhance its versatility.

    \item \textbf{Interpretable AI}: 
    By making the relational reasoning in attention explicit, these methods can pave the way for more interpretable AI models, aiding domains where transparency is critical, such as healthcare, environmental studies, and scientific discovery.

    \item \textbf{Efficient Fine-Tuning Techniques}: 
    Sparse GNN fine-tuning, with its minimal computational footprint, offers a scalable alternative for adapting large pre-trained models to specialized tasks. Exploring its applications in low-resource settings or edge computing could be impactful, along with additional experimentation to test model performance in real-world applications like materiomic reasoning.

    \item \textbf{Advanced Graph Reasoning}:
    Building on the theoretical insights, integrating more complex graph reasoning techniques, such as deeper message passing neural networks or hierarchical graph structures, could yield even more expressive models.

    \item \textbf{Foundational AI Research}:
    The reinterpretation of Transformers as hierarchical GIN models may be a nucleus for deeper exploration into their theoretical underpinnings. This perspective could lead to new architectures that unify graph neural networks and Transformers into a cohesive framework.

    \item \textbf{Application to Intersections between Natural and Social Systems}:
    Beyond technical advancements, these methods hold potential for studying natural systems (e.g., protein folding or ecological networks) and social systems (e.g., social media dynamics or economic modeling), where relational structures are paramount, and where universal patterns could be critical to achieve generalization and the extraction of new knowledge.
\end{enumerate}

It may be worthwhile to explore how the focus on graphs may also help develop not only better performing models but also provide pathways for improving reasoning models that focus on inference compute scaling~\cite{o1-model-card-2024,plaat2024reasoninglargelanguagemodels,buehler2024preflexorpreferencebasedrecursivelanguage,zelikman2022starbootstrappingreasoningreasoning,zelikman2024quietstarlanguagemodelsteach,wei2023chainofthoughtpromptingelicitsreasoning,kojima2023largelanguagemodelszeroshot,mischler2024contextual,li2024chainthoughtempowerstransformers}. For instance, we could feed relations modeled by the GIN-Attention mechanisms (or Sparse-GIN) back into the reasoning process and train models with increased mechanistic awareness. This additional signal could serve as powerful cues in the development of reasoning models. A specific idea is to utilize graph representations, perhaps processed or sparsified relationships, to inform, train and evolve the reasoning process in hidden dimensions without ever decoding to discrete tokens. Such self-reflection in embedding space had been explored in earlier work~\cite{Buehler2024X-LoRA:Design,buehler2024preflexorpreferencebasedrecursivelanguage} and could be a strategy to inference-scale performance with relevant reward signals during fine-tuning.  

By synthesizing graph-based reasoning with the proven capabilities of Transformers, we may build a new generation of models that are more flexible, interpretable, and capable of tackling complex, real-world challenges. While much further research is necessary, this work offers a new perspective on attention in Transformer architectures by integrating graph isomorphic reasoning and dynamic graph-based fine-tuning, enabling unprecedented adaptability, generalization, and interpretability for tackling complex relational data across diverse domains. The ease of adapting sparse GIN models into any existing Transformer architecture may be a particularly appealing way to explore such opportunities.

\begin{figure}
	\centering
	\includegraphics[width=1.\linewidth]{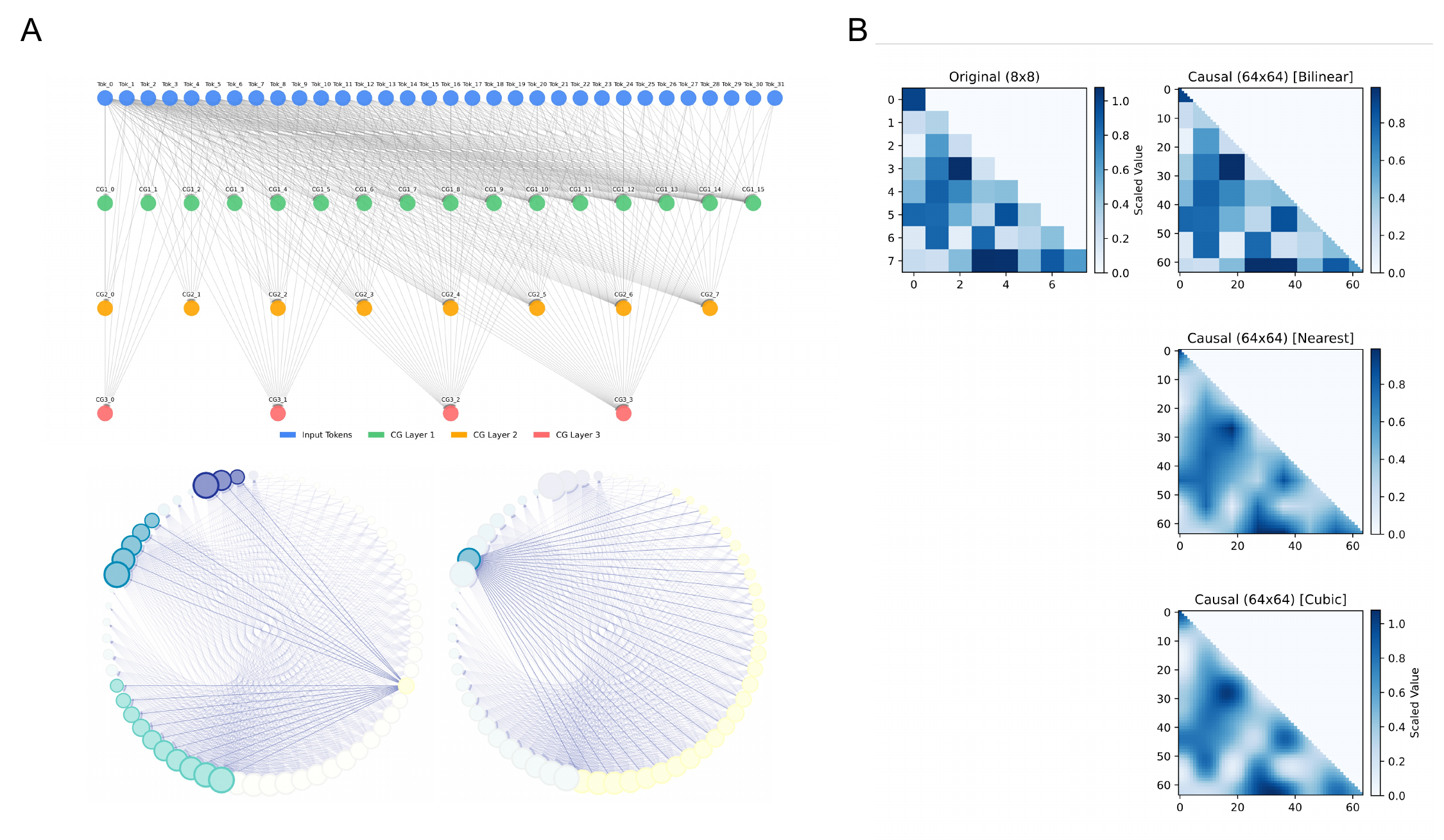}
        \caption{Exploration of future alternative graph-based attention and Transformer models that engage heavily with coarse-grained representations. Panel A: Tokens are mapped into different levels of coarser representations, in which then graph operations are conducted. For decoding, one can either find suitable methods to decode coarse tokens (in a way that respects causality) or decode tokens at the fine level during inference and task solution as done in~\cite{hawthorne2022generalpurposelongcontextautoregressivemodeling}. Panel B: Scaled coarse-grained attention, where we simplify the computation of adjacency matrices by conducting their construction in a coarse representation space. For computing graph operations, these coarse adjacency are upscaled (using bilinear, nearest cubic, etc. algorithms) to the full-scale resolution, where they are used to conduct graph operations. The advantage of the latter approach is that it remains fully causal. }
	\label{fig:fig_3000}
\end{figure}

Other opportunities may exist when graph reasoning in attention as explored in this paper is combined with coarse-grain (CG) modeling, a method widely used in multiscale simulation of physical and other systems~\cite{Tarakanova2019Coarse-grainedFibrils,Wang2019MachineFields,Cranford2012Biomateriomics,Yeo2018MultiscaleSimulations}. Thereby, groups of tokens could be mapped to coarser representations and further strengthen capabilities in hiearchical reasoning. Figure~\ref{fig:fig_3000} shows two possible scenarios. Figure~\ref{fig:fig_3000}A visualizes a method by which tokens are mapped into different levels of coarser representations, in which  graph operations are then conducted. For decoding to higher-resolution space one can either find suitable methods to decode coarse tokens (in a way that respects causality) or decode tokens at the fine level during inference and task solution as proposed in~\cite{hawthorne2022generalpurposelongcontextautoregressivemodeling}. Recent related work such as the use of Byte Latent Transformer models~\cite{pagnoni2024bytelatenttransformerpatches} represent an effective coarse-grained representation is learned at byte-level. In another approach outlined in Figure~\ref{fig:fig_3000}B, we propose scaled coarse-graine attention, which may simplify the computation of adjacency matrices for graph reasoning. The core idea is to conduct adjacency matrices for graph reasoning in a learned coarse representation space, taking advantage of related or concerted interaction between tokens. For computing graph operations at the fine level, the coarse-level adjacency matrices can be upscaled (using bilinear, nearest cubic, etc. algorithms) to the full-scale resolution, where they are used to conduct graph operations. The advantage of the latter approach is that it remains easily fully causal since all attention computations are done in high-resolution space. 

\section{Materials and Methods}

Much of the theoretical and methodological details have been provided in the main paper. Here we include additional aspects and practical steps. We refer to the GitHub repository source code for additional specifics in the algorithmic steps. 

\subsection{Graph-aware attention: Training task, custom tokenizer and model construction}

To assess performance of the graph-aware attention approaches, we develop and train a custom model with the same training set from scratch. We use \url{lamm-mit/protein_secondary_structure_from_PDB}~\cite{Yu2022End-to-EndProteins} for all Protein Data Bank (PDB) proteins (\url{https://www.rcsb.org/}) with up to 256 amino acid sequence length, creating a task to dominating secondary structure features based on the sequence. We predict both the overall dominating secondary structure type and second-ranked secondary structure type, as follows:
    
\begin{LLMbox}{}
\begin{lstlisting}
<|begin_of_text|><|sequence|>[SEQUENCE]<|/sequence|><|Primary_SS_Type|><|Secondary_SS_Type}|><|eot_id|>
\end{lstlisting}
\end{LLMbox}

Samples from the training set are:

\begin{LLMbox}{Data set sample}
\begin{lstlisting}
<|begin_of_text|><|sequence|>PLIVPYNLPLPGGVV...LNEISKLGISGDIDLTSASYTMI<|/sequence|><|BS|><|UNSTRUCTURED|><|eot_id|>
<|begin_of_text|><|sequence|>ADDIVFKAKNGDVKFPHKA...GCHEEMKKGPTKCGECHKK<|/sequence|><|T|><|T|><|eot_id|>
<|begin_of_text|><|sequence|>SLQDPFLNALRRERVPVSI...QMVYKHAISTVVPSRPVSH<|/sequence|><|BS|><|UNSTRUCTURED|><|eot_id|>
<|begin_of_text|><|sequence|>SMEQVAMELRLTELTRLLRSVLD...SIGLE<|/sequence|><|AH|><|UNSTRUCTURED|><|eot_id|>
\end{lstlisting}
\end{LLMbox}

We split the dataset into 90\% training and 10\% test data. 

We train a custom tokenizer with special tokens for each of the amino acids, secondary structures and others:

\begin{LLMbox}{Special tokens}
\begin{lstlisting}
# Special tokens
"<|pad|>",
"<|eot_id|>", 
"<|begin_of_text|>",
"<|unk|>",
"<|mask|>",
"<|sequence|>",
"<|/sequence|>"
# Single-letter amino acid codes
"A", "R", "N", "D", "C", "E", "Q", "G", "H", "I", "L", "K", "M", "F", "P", "S", "T", "W", "Y", "V"
# Additional special words
"<|AH|>", "<|BS|>", "<|UNSTRUCTURED|>", "<|BEND|>", "<|PHIHELIX|>", "<|310HELIX|>", "<|BETABRIDGE|>", "<|T|>"
\end{lstlisting}
\end{LLMbox}

The tokenizer has a vocabulary size of 271 unique tokens. 

The model architectures all feature $N_x=6$ layers, and 8 heads each, following the Llama 3 architecture and base parameters (like dropout, RMS norm, etc.) (as introduced in \texttt{meta-llama/Meta-Llama-3-8B-Instruct}). The model all feature the same standard feed-forward MLP and other architectural components, except for the attention module that is constructed differently. The hidden dimension in is adjusted in each model so that the resulting final parameter count is around 25 M parameter in either model choice. For instance, in the best performing GIN-Attention model the hidden dimension is 528, whereas it is 512 in the standard transformer variant (the GIN-Attention case still performs better than the reference case even if the hidden dimension is identical, even though it has fewer parameters in that case). We use key-value grouping equal to one in all cases. We experimented also with a case where we only adapted the hidden dimension in the attention mechanism but left the inner, expanded dimension in the Transformer FF MLP the same. This case, along with removing the o\_proj layer after the GIN MLP, yielded the best overall performance. 

A summary of other training and hyperparameters is provided in Table~\ref{tab:key_hyperparameters_GIN}.

\begin{table}[h!]
\small
\centering
\begin{tabular}{|c|l|}
\hline
\textbf{Hyperparameter} & \textbf{Value} \\ \hline
Learning Rate & \(1 \times 10^{-4}\) \\ \hline
Per Device Train Batch Size & 8 \\ \hline
Per Device Eval Batch Size & 4 \\ \hline
Gradient Accumulation Steps & 4 \\ \hline
Number of Training Epochs & 9 \\ \hline
Weight Decay & 0.01 \\ \hline
Learning Rate Scheduler Type & Constant \\ \hline
Warmup Steps & 250 \\ \hline
Packing & False \\ \hline
Max Gradient Norm & 1 \\ \hline
\end{tabular}
\caption{Hyperparameters for model training (reference, GIN-Attention, PNA-Attention, and all variants), implemented in SFTTrainer that is part of the Hugging Face Transformer Reinforcement Learning (TRL) package (\url{https://huggingface.co/docs/trl/en/index}). Train and test loss is computed every 100 steps.}
\label{tab:key_hyperparameters_GIN}
\end{table}

\subsection{Sparse GIN fine-tuning model}

We use the \texttt{mlabonne/orca-math-word-problems-80k} dataset~\cite{mitra2024orcamathunlockingpotentialslms} for training, and create instruction prompts using this format:

\begin{LLMbox}{Data set formatting}
\begin{lstlisting}
### User: [QUESTION]<|eot_id|>### Assistant: [ANSWER]<|eot_id|>" 
\end{lstlisting}
\end{LLMbox}

One example is:

\begin{LLMbox}{Data set example}
\begin{lstlisting}
### User: Macey saves to buy herself a shirt. She was able to save $1.50 already. She needs to save for 3 more weeks at $0.50 per week to have enough for the shirt. How much does the shirt cost?<|eot_id|>### Assistant: Macey has already saved $1.50. She plans to save for 3 more weeks at a rate of $0.50 per week. \n\nThe total amount she will save in the next 3 weeks is:\n3 weeks * $0.50/week = $1.50\n\nAdding this to the amount she has already saved:\n$1.50 (already saved) + $1.50 (to be saved) = $3.00\n\nTherefore, the shirt costs $3.00.<|eot_id|>'
\end{lstlisting}
\end{LLMbox}

We split the dataset into 90\% training and 10\% test data. 

\begin{figure}
	\centering
	\includegraphics[width=1.\linewidth]{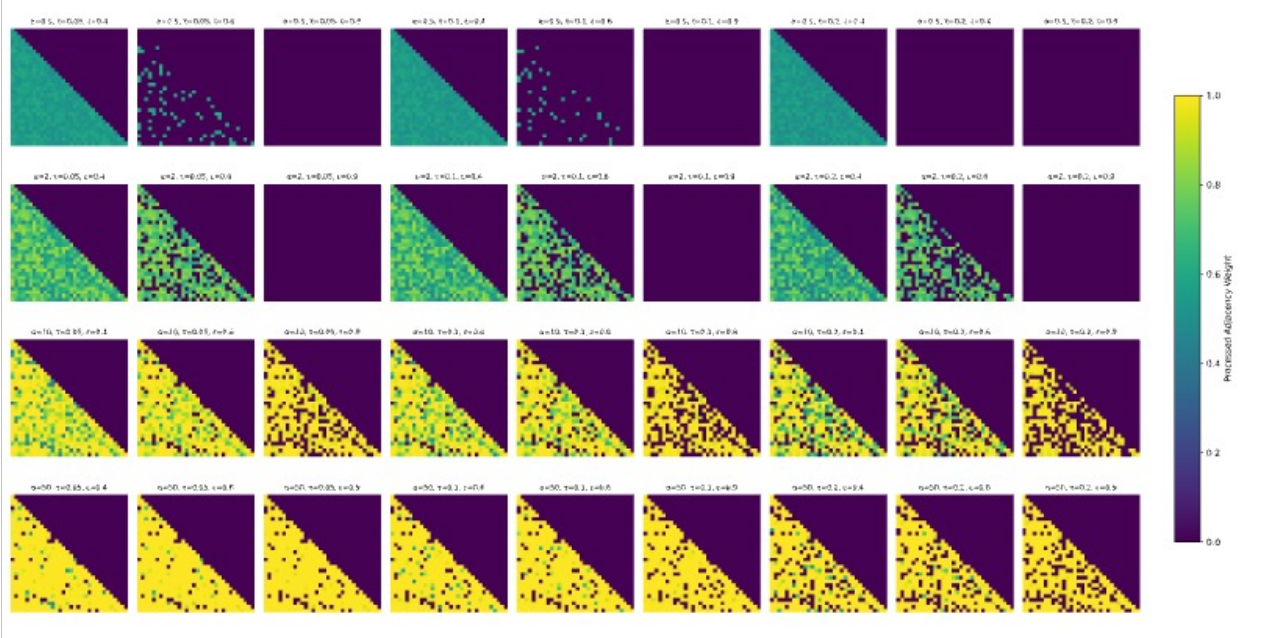}
        \caption{Sharpening and thresholding techniques implemented to sparsify the attention matrix for use as fine-tuning algorithm. This plot shows the results of sharpening ($\tau$ and $\alpha$) and thresholding ($\varepsilon$). As one can see, sharpening yields sharper distributions, which are then sparsified using the discrete threshold to yield close to binary \hlyellow{adjacency} matrices. As reference, for the Sparse GIN fine-tuning model, we use $\tau=0.1$, a sharpening value of $\alpha=10.0$, and a sparsification threshold of $\varepsilon=0.6$.}
	\label{fig:fig_2000}
\end{figure}

As pre-trained model we use the \texttt{meta-llama/Llama-3.2-3B-Instruct} foundation model. For the LoRA variant, we create adapters for \texttt{q\_proj}, \texttt{k\_proj}, \texttt{v\_proj}, and \texttt{o\_proj} with rank $r=50$ and $\alpha=50$. For the Sparse GIN model, we use $\tau=0.1$ and a sharpening value of $\alpha=10.0$. We use a sparsification threshold of $\varepsilon=0.6$. These parameters are not trainable in our experiment. The effect of the various processing techniques implemented are shown in Figure~\ref{fig:fig_2000}, for a randomly generated causal set of values. The hidden dimension of the GIN layers is 155. Both variants have approximately 28 M trainable parameters (the pre-trained model has around 3B parameters, which are frozen). 

A set of other training and hyperparameters is summarized in Table~\ref{tab:key_hyperparameters_Sparse_GIN}.

\begin{table}[h!]
\small
\centering
\begin{tabular}{|c|l|}
\hline
\textbf{Hyperparameter} & \textbf{Value} \\ \hline
Learning Rate & \(2 \times 10^{-4}\) \\ \hline
Per Device Train Batch Size & 1 \\ \hline
Per Device Eval Batch Size & 2 \\ \hline
Gradient Accumulation Steps & 4 \\ \hline
Number of Training Epochs & 3 \\ \hline
Weight Decay & 0.01 \\ \hline
Learning Rate Scheduler Type & Constant \\ \hline
Warmup Steps & 50 \\ \hline
Packing & False \\ \hline
Max Gradient Norm & 0.5 \\ \hline
\end{tabular}
\caption{Hyperparameters for LoRA and Sparse-GIN model training, implemented in SFTTrainer (\url{https://huggingface.co/docs/trl/en/index}).}
\label{tab:key_hyperparameters_Sparse_GIN}
\end{table}

In the second example, we add LoRA layers in all linear layers in the model except for embedding and head layer. The model is trained with the \url{lamm-mit/SAGI-1-SYMBOLIC_DATA_PLUS_REASONING_DATA_V1_100K} dataset, consisting of 100K logic and reasoning questions. 

We further trained a Sparse-GIN model on the \url{lamm-mit/bio-silk-mech-mix-q-a-35K} dataset.

\begin{LLMbox}{Data set example \# 1}
\begin{lstlisting}
### User: Even if a crime that has been committed by computer is discovered and reported, the odds of being both arrested and convicted greatly favor the criminal. Each of the following, if true, supports the claim above EXCEPT: A: The priorities of local police departments, under whose jurisdiction most computer crime falls, are weighted toward visible street crime that communities perceive as threatening. B: Computer criminals have rarely been sentenced to serve time in prison, because prisons are overcrowded with violent criminals and drug offenders. C: The preparation of computer-fraud cases takes much more time than is required for average fraud cases, and the productivity of prosecutors is evaluated by the number of good cases made. D: The many police officers who are untrained in computers often inadvertently destroy the physical evidence of computer crime.<|eot_id|>### Assistant: B<|eot_id|>
\end{lstlisting}
\end{LLMbox}

\begin{LLMbox}{Data set example \# 2}
\begin{lstlisting}
### User: Why would a person go to school?<|eot_id|>### Assistant: Define: Person(x): x is a person GoToSchool(x): x goes to school SuccessfulInLife(x): x is successful in life Given: (*@$\forall$@*) x(Person(x) (*@$\rightarrow$@*) (GoToSchool(x) (*@$\rightarrow$@*) SuccessfulInLife(x))) Implies: Person(x) (*@$\rightarrow$@*) (GoToSchool(x) (*@$\rightarrow$@*) SuccessfulInLife(x)) Explanation: This statement implies that for all values of 'x', if 'x' is a person, then if 'x' goes to school, they will be successful in life. Therefore, a person goes to school in order to be successful in life.
\end{lstlisting}
\end{LLMbox}

\section*{Code and data}

All codes and data are available at \url{https://github.com/lamm-mit/Graph-Aware-Transformers}.

\section*{Acknowledgments}

The author acknowledges support from the MIT Generative AI initiative.

\bibliographystyle{naturemag}

\bibliography{references,references-Mendeley}  

\end{document}